\documentclass{article}

\usepackage{microtype}
\usepackage{graphicx}
\usepackage{subfigure}
\usepackage{booktabs} 

\usepackage{hyperref}

\usepackage[accepted]{icml2021}

\usepackage[utf8]{inputenc}
\usepackage{amsmath}
\usepackage{amsfonts}
\usepackage{amssymb}
\usepackage{amsthm}
\usepackage{graphicx}
\usepackage{dsfont}
\usepackage{algorithm}
\usepackage{algorithmic}
\usepackage{bbm}
\usepackage{multirow}

\newtheorem{theorem}{Theorem}
\newtheorem{proposition}[theorem]{Proposition}
\newtheorem{lemma}[theorem]{Lemma}
\newtheorem{corollary}[theorem]{Corollary}
\newtheorem*{remark}{Remark}

\newtheorem{innercustomgeneric}{\customgenericname}
\providecommand{\customgenericname}{}
\newcommand{\newcustomtheorem}[2]{%
	\newenvironment{#1}[1]
	{%
		\renewcommand\customgenericname{#2}%
		\renewcommand\theinnercustomgeneric{##1}%
		\innercustomgeneric
	}
	{\endinnercustomgeneric}
}
\newcustomtheorem{customtheorem}{Theorem}
\newcustomtheorem{customproposition}{Proposition}
\newcustomtheorem{customlemma}{Lemma}
\newcustomtheorem{customcorollary}{Corollary}

\DeclareMathOperator*{\argmax}{arg\,max}

\newcommand{\KL}{\mathbb{KL}}
\newcommand{\E}{\mathbb{E}}
\newcommand{\D}{\mathcal{D}}

\newcommand{\R}{\mathbb{R}}

\newcommand{\B}{\mathcal{B}}
\newcommand{\T}{\mathcal{T}}

\newcommand{\1}{\mathbbm{1}}

\renewcommand{\algorithmicrequire}{\textbf{Input:}}
\renewcommand{\algorithmicensure}{\textbf{Output:}}
\renewcommand{\P}{\Delta}
\renewcommand{\S}{S}

\newcommand{\A}{A}
\newcommand{\SA}{{\S\times\A}}

\newcommand{\propositionref}[1]{\textbf{Proposition~\ref{#1}}}
\newcommand{\figureref}[1]{\textit{Figure~\ref{#1}}}

\newcommand{\piD}{\pi_{\!_D}}

\definecolor{green}{rgb}{0,0,0}

\icmltitlerunning{OptiDICE: Offline Policy Optimization via Stationary Distribution Correction Estimation}

\begin{document}

\twocolumn[
\icmltitle{OptiDICE: Offline Policy Optimization via \\ Stationary Distribution Correction Estimation}



\icmlsetsymbol{equal}{*}

\begin{icmlauthorlist}
\icmlauthor{Jongmin Lee}{kaistcs,equal}
\icmlauthor{Wonseok Jeon}{mila,mcgill,equal}
\icmlauthor{Byung-Jun Lee}{gausslabs}
\icmlauthor{Joelle Pineau}{mila,mcgill,fair}
\icmlauthor{Kee-Eung Kim}{kaistcs,kaistai}
\end{icmlauthorlist}

\icmlaffiliation{kaistcs}{School of Computing, KAIST}
\icmlaffiliation{kaistai}{Graduate School of AI, KAIST}
\icmlaffiliation{mila}{Mila, Quebec AI Institute}
\icmlaffiliation{mcgill}{School of Computer Science, McGill University}
\icmlaffiliation{fair}{Facebook AI Research}
\icmlaffiliation{gausslabs}{Gauss Labs Inc.}

\icmlcorrespondingauthor{Jongmin Lee}{jmlee@ai.kaist.ac.kr}
\icmlcorrespondingauthor{Wonseok Jeon}{jeonwons@mila.quebec}

\icmlkeywords{Machine Learning, ICML}

\vskip 0.3in
]



\printAffiliationsAndNotice{\icmlEqualContribution} 

\begin{abstract}
We consider the offline reinforcement learning (RL) setting where the agent aims to optimize the policy solely from the data without further environment interactions.
In offline RL, the distributional shift becomes the primary source of difficulty,
which arises from the deviation of the target policy being optimized from the
behavior policy used for data collection. This typically causes overestimation of
action values, which poses severe problems for model-free algorithms 
that use bootstrapping. 
To mitigate the problem, prior offline RL algorithms often 
used sophisticated techniques that encourage underestimation of action values,
which introduces an additional set of hyperparameters that need to be tuned properly. 
In this paper, we present an offline RL algorithm that prevents overestimation
in a more principled way.
Our algorithm, OptiDICE, directly estimates the stationary distribution corrections of the optimal policy and does not rely on policy-gradients, unlike previous offline RL algorithms.
Using an extensive set of 
benchmark datasets for offline RL, we show that OptiDICE performs competitively
with the state-of-the-art methods.

\end{abstract}

\section{Introduction}

The availability of large-scale datasets has been one of the important
factors contributing to the recent success in machine learning 
for real-world tasks such as computer 
vision~\cite{deng2009imagenet,alex2012imagenet} and natural 
language processing~\cite{devlin2019bert}. The standard workflow 
in developing systems for typical machine learning tasks is 
to train and validate the model 
on the dataset, and then to deploy the model with its parameter fixed 
when we are satisfied with training. 
This offline training allows us to address various operational requirements of the system
without actual deployment, such as acceptable level of prediction accuracy rate
once the system goes online.


However, this workflow is not straightforwardly applicable to 
the standard setting of reinforcement learning (RL)~\cite{sutton1998rlbook}
because of the online learning assumption: the RL agent needs to 
continuously explore the environment and learn from its trial-and-error 
experiences 
to be properly trained.
This aspect has been one of the fundamental bottlenecks for the 
practical adoption of RL in many real-world domains, where the
exploratory behaviors are costly or even dangerous, 
e.g. autonomous driving~\cite{yu2020bdd100k} and clinical treatment~\cite{yu2020reinforcement}.


Offline RL (also referred to as batch RL) \cite{ernst2005tree,lange2012,fujimoto2019off,levine2020offline}  
casts the RL problem in the offline training setting. One of the most relevant areas of
research in this regard is the off-policy RL~\cite{lillicrap2016,haarnoja2018soft,fujimoto2018addressing}, since we
need to deal with the distributional shift resulting from the trained policy being deviated from the policy used to collect
the data. However, without the data continuously collected online, this distributional 
shift cannot be reliably corrected and poses a significant challenge to
RL algorithms that employ bootstrapping together with function approximation:
it causes compounding overestimation of the action values for model-free 
algorithms~\cite{fujimoto2019off,kumar2019stabilizing}, which arises 
from computing the bootstrapped target using the predicted values of 
\emph{out-of-distribution} actions. 
To mitigate the problem, most of the current 
offline RL algorithms have proposed 
sophisticated techniques to encourage underestimation of action values,
introducing an additional set of hyperparameters that needs to be tuned properly \cite{fujimoto2019off,kumar2019stabilizing,jaques2019way,lee2020batch,kumar2020conservative}.


In this paper, we present an offline RL algorithm that essentially 
eliminates the need to evaluate out-of-distribution actions, thus avoiding
the problematic overestimation of values. 
Our algorithm, \emph{Offline Policy \textbf{Opti}mization via Stationary \textbf{DI}stribution \textbf{C}orrection \textbf{E}stimation} (OptiDICE),
estimates stationary distribution ratios that correct the discrepancy between the data distribution and the \emph{optimal} policy's stationary distribution.
We first show that such optimal stationary distribution corrections can be estimated via  minimax optimization that does not involve 
sampling from the target policy.
Then, we derive and exploit the closed-form solution to the sub-problem of the aforementioned minimax optimization, which reduces the overall problem into an unconstrained convex optimization, and thus greatly stabilizing our method.
To the best of our knowledge, OptiDICE is the first deep offline RL algorithm that optimizes policy purely in the space of \emph{stationary distributions}, rather than in the space of either Q-functions or policies \cite{nachum2019algaedice}.
In the experiments, we demonstrate that OptiDICE performs competitively with the state-of-the-art methods using the D4RL offline RL benchmarks \cite{fu2020d4rl}.

\section{Background}

We consider the reinforcement learning problem with the environment modeled as a Markov Decision Process (MDP)
$M = \langle \S, \A, T, R, p_0, \gamma\rangle$~\cite{sutton1998rlbook}, where $S$ is the set of states $s$, $A$ is the set of actions $a$, $R: \S \times \A \rightarrow \R$ is the reward function, $T: \S \times \A \rightarrow \P(\S)$ is a transition probability, $p_0\in\P(\S)$ is an initial state distribution, and $\gamma \in [0, 1]$ is a discount factor. The policy $\pi: \S \rightarrow \P(\A)$ is a mapping from state to distribution over actions. 
While $T(s, a)$ and $\pi(s)$ indicate distributions by definition, we let $T(s'|s, a)$ and $\pi(a|s)$ denote their evaluations for brevity.
For the given policy $\pi$, the stationary distribution $d^\pi$ is defined as
\begin{align*}
d^\pi(s,a)
= \begin{cases}
    (1 - \gamma) \sum\limits_{t=0}^\infty \gamma^t \Pr(s_t = s, a_t = a) & \hspace{-2pt}\text{if } \gamma < 1, \\
    \lim\limits_{T \rightarrow \infty} \frac{1}{T + 1} \sum\limits_{t=0}^T \Pr(s_t = s, a_t = a) & \hspace{-2pt}\text{if } \gamma = 1,
\end{cases}
\end{align*}
where $s_0 \sim p_0$ and $a_t \sim \pi(s_t), s_{t+1} \sim T(s_t, a_t)$ for all time step $t$. 
The goal of RL is to learn an optimal policy that maximizes rewards through interactions with the environment: 
$\max_\pi \E_{(s,a) \sim d^\pi} [ R(s,a) ]$. The value functions of policy $\pi$ is defined as $Q^\pi(s,a):= \E_{\pi, M} \left[ \sum_{t=0}^\infty \gamma^t R(s_t, a_t) | s_0 = s, a_0 = a \right]$ and $V^\pi(s):=\E_{a\sim \pi(s)}[Q^\pi(s, a)]$, where the action-value function $Q^\pi$ is a unique solution of the Bellman equation:
\begin{align*}
    Q^\pi(s,a) = R(s,a) + \gamma \E_{\substack{s' \sim T(s,a) \\ a' \sim \pi(s')}} [ Q^\pi(s', a') ].
\end{align*}

In offline RL, the agent optimizes the policy from static dataset $D = \{(s_i, a_i, r_i, s_i')\}_{i=1}^N$ 
collected before the training phase.
We denote the empirical distribution of the dataset by $d^D$ and will abuse the notation $d^D$ to represent $s\sim d^D$, $(s,a) \sim d^D$, and $(s,a,s') \sim d^D$.

Prior offline model-free RL algorithms, exemplified by \cite{fujimoto2019off,kumar2019stabilizing,wu2019behavior,lee2020batch,kumar2020conservative,nachum2019algaedice}, rely on estimating Q-values for optimizing the target policy.
This procedure often yields unreasonably high Q-values due to
the compounding error from bootstrapped estimation with out-of-distribution actions sampled from the target policy \cite{kumar2019stabilizing}.

\section{OptiDICE}
In this section, we present \emph{Offline Policy \textbf{Opti}mization via Stationary \textbf{DI}stribution \textbf{C}orrection \textbf{E}stimation} (OptiDICE).
Instead of the \emph{optimism in the face of uncertainty} principle \cite{szita2008the} in online RL, we
discourage the uncertainty as in most offline RL algorithms~\cite{kidambi2020morel,yu2020mopo}; otherwise, the resulting policy may fail to improve on the data-collection policy, or even suffer from severe performance degradation \cite{petrik2016safe,laroche2019safe}.
Specifically, we consider the regularized policy optimization framework \cite{nachum2019algaedice}
\begin{align}
    \pi^* := \argmax_{\pi}
    \E_{(s, a)\sim d^\pi}[R(s, a)]
    -
    \alpha D_f(d^\pi||d^D),
    \label{eq:objective_pi}
\end{align}
where $D_f(d^\pi||d^D):=\E_{(s,a) \sim d^D} \left[f\big(\frac{d^\pi(s,a)}{d^D(s,a)}\big) \right]$ is the $f$-divergence between the stationary distribution $d^\pi$ and the dataset distribution $d^D$,
and $\alpha > 0$ is a hyperparameter that balances between pursuing the reward-maximization and penalizing the deviation from the distribution of the offline dataset (i.e. penalizing distributional shift).
We assume $d^D>0$ and $f$ being strictly convex and continuously differentiable. 
Note that we impose regularization in the space of stationary distributions rather than in the space of policies \cite{wu2019behavior}.
However, optimizing for $\pi$ in \eqref{eq:objective_pi} involves the evaluation of $d^\pi$, which is not directly accessible in the offline RL setting.

To make the optimization tractable, we reformulate \eqref{eq:objective_pi} in terms of optimizing a stationary distribution $d:\SA\rightarrow\R$. For brevity, we consider discounted MDPs ($\gamma < 1$) and then generalize the result to undiscounted MDPs ($\gamma = 1$). Using $d$, we rewrite \eqref{eq:objective_pi} as
\begin{align}
    \max_{d}~
    &
    \E_{(s, a)\sim d} [R(s, a)] - \alpha D_f(d||d^D) \label{eq:objective_d} \\
    \mathrm{s.t.}~
    & (\B_* d)(s) = (1 - \gamma) p_0(s) + \gamma (\T_* d)(s) ~~ \forall s, \label{eq:objective_d_constraint1} 
    \\
	& d(s, a)\ge0 ~~ \forall s, a, \label{eq:objective_d_constraint2}
\end{align}
where $(\B_*d)(s) := \sum_{\bar a} d(s, \bar a)$ is a marginalization operator, and $(\T_*d)(s) := \sum_{\bar s, \bar a} T(s | \bar s, \bar a) d(\bar s, \bar a)$ is a transposed Bellman operator\footnote{
While AlgaeDICE \cite{nachum2019algaedice} also 
proposes $f$-divergence-regularized policy optimization as \eqref{eq:objective_pi}, it imposes
Bellman flow constraints on state-action pairs, whereas our formulation imposes constraints only on
states, which is more natural for finding the
optimal policy.
}.
Note that when $\alpha=0$, the optimization~(\ref{eq:objective_d}-\ref{eq:objective_d_constraint2}) is exactly the dual formulation of the linear program (LP) for finding an optimal policy of the MDP \cite{puterman1994markov}, where the constraints (\ref{eq:objective_d_constraint1}-\ref{eq:objective_d_constraint2}) are often called the Bellman flow constraints.
Once the optimal stationary distribution $d^*$ is obtained, we can recover the optimal policy $\pi^*$ in \eqref{eq:objective_pi}  from $d^*$ by $\pi^*(a|s) = \frac{ d^*(s,a) }{\sum_{\bar a} d^*(s, \bar a)}$.

We then obtain the following Lagrangian for the
constrained optimization problem in (\ref{eq:objective_d}-\ref{eq:objective_d_constraint2}):
\begin{align}
    &\max_{d \ge 0} \min_\nu 
    \E_{(s, a)\sim d}[R(s, a)] - \alpha D_f(d||d^D) \label{eq:objective_lagrangian_d} \\
    &~~~~~~+
    \textstyle\sum\limits_{s} \nu(s)
    \Big( (1 - \gamma) p_0(s) + \gamma (\T_* d)(s) - (\B_* d)(s) \Big), \nonumber
\end{align}
where $\nu(s)$ are the Lagrange multipliers.
Lastly, we eliminate the direct dependence on $d$ 
and $\mathcal{T}_*$ by rearranging the terms in~\eqref{eq:objective_lagrangian_d} and 
optimizing the distribution ratio $w$ instead of
$d$:
\begin{align}
    &\hspace{-2ex}\E_{(s, a)\sim d}[R(s, a)] - \alpha D_f(d||d^D) \nonumber \\
    &+ \textstyle\sum\limits_{s} \nu(s) \Big( (1 - \gamma) p_0(s) + \gamma (\T_* d)(s) - (\B_* d)(s) \Big) \nonumber \\
    =& \textstyle (1 - \gamma) \E_{s \sim p_0}[\nu(s)] + \E_{(s,a) \sim d^D} \hspace{-2pt} \left[ - \alpha f \left( \frac{d(s,a)}{d^D(s,a)} \right) \right] \label{eq:objective_lagrangian_d2} \\
    &+ \textstyle\sum\limits_{s,a} d(s,a) \Big( \hspace{-3pt}\underbrace{R(s,a) + \gamma (\T \nu)(s,a) - (\B \nu)(s,a)}_{
    \let\scriptstyle\textstyle \small \substack{
    =: e_\nu(s,a) \text{~~(`advantage' using $\nu$)}
    }} \hspace{-3pt} \Big) \nonumber \\
    =& \textstyle (1 - \gamma) \E_{s \sim p_0}[\nu(s)] + \E_{(s,a) \sim d^D}\left[ - \alpha f \left( \frac{d(s,a)}{d^D(s,a)} \right) \right]  \nonumber \\
    & \textstyle+ \E_{(s,a) \sim d^D} \Big[ \hspace{-4pt} \underbrace{\tfrac{d(s,a)}{d^D(s,a)}}_{
    \let\scriptstyle\textstyle \small \substack{
    =: w(s,a)
    }} \hspace{-4pt} \big( e_\nu(s,a) \big) \Big] \nonumber \\
    =& (1 - \gamma) \E_{s \sim p_0}[\nu(s)] + \E_{(s,a) \sim d^D}\left[-\alpha f \big( w(s,a) \big) \right] \nonumber \\
    &+ \E_{(s,a) \sim d^D} \left[ w(s,a) \big( e_\nu(s,a) \big) \right]
    =: L(w, \nu).
    \label{eq:objective_lagrangian_w}
\end{align}
The first equality holds due to the property of the adjoint (transpose) operators $\B_*$ and $\T_*$, i.e. for any $\nu$,
\begin{align*}
    \textstyle\sum\limits_s \nu(s)(\B_* d)(s)
    &=
    \textstyle\sum\limits_{s, a} d(s, a) (\B \nu)(s, a),\\
    \textstyle\sum\limits_s \nu(s)(\T_* d)(s)
    &=
    \textstyle\sum\limits_{s, a} d(s, a) (\T \nu)(s, a),
\end{align*}
where $(\T \nu)(s, a) = \sum_{s'} T(s' | s,a) \nu(s')$ and $(\B \nu)(s,a) = \nu(s)$.
Note that $L(w, \nu)$ in \eqref{eq:objective_lagrangian_w} does not involve expectation over $d$, but only expectation over $p_0$ and $d^D$, which allows us to perform 
optimization only with the offline data.
\begin{remark}
The terms 
in~\eqref{eq:objective_lagrangian_w} will be estimated only by using the samples from the dataset distribution $d^D$:
\begin{align}
    &\hat L(w, \nu) := (1 - \gamma) \E_{s \sim p_0}[\nu(s)]  \label{eq:objective_minmax_sars1} \\
    &+ \E_{(s,a,s') \sim d^D} \left[- \alpha f \big( w(s,a) \big) + w(s,a) \big( \hat e_\nu(s,a,s') \big) \right]. \nonumber
\end{align}
Here, $\hat e_\nu(s,a,s') := R(s,a) + \gamma \nu(s') - \nu(s)$ is a single-sample estimation of advantage $e_\nu(s, a)$.
On the other hand, prior offline RL algorithms often involve  
estimations using out-of-distribution actions sampled from the target policy, e.g. employing a critic to compute
bootstrapped targets for the value function. Thus, 
our method is free from the compounding
error in the bootstrapped estimation due to 
using out-of-distribution actions.
\end{remark}

In short,
OptiDICE solves the problem
\begin{align}
    \max_{w \ge 0} \min_{\nu} L(w, \nu),
    \label{eq:L_maxmin}
\end{align}
where the optimal solution $w^*$  of the optimization~\eqref{eq:L_maxmin} represents the stationary distribution corrections between the \emph{optimal} policy's stationary distribution and the dataset distribution:
$w^*(s,a) = \frac{d^{\pi^*}(s,a)}{d^D(s,a)}$.

\begin{figure*}[t!]
\centering     
\subfigure[Stationary Dist. $d^{\piD}$]{\label{figure:illustrative_example:a}\includegraphics[width=0.49\columnwidth]{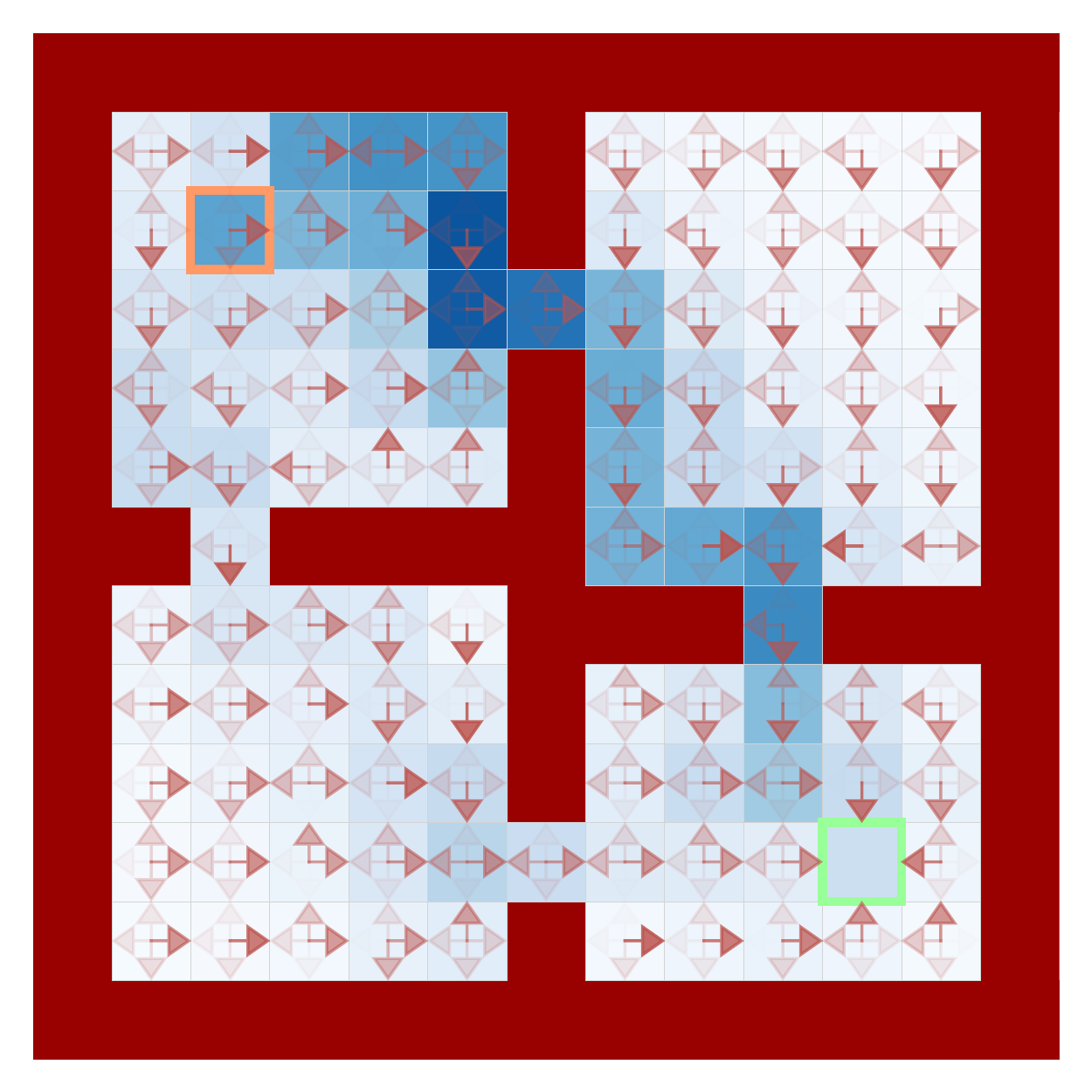}}
\subfigure[Empirical Dist. $d^D$]{\label{figure:illustrative_example:b}\includegraphics[width=0.49\columnwidth]{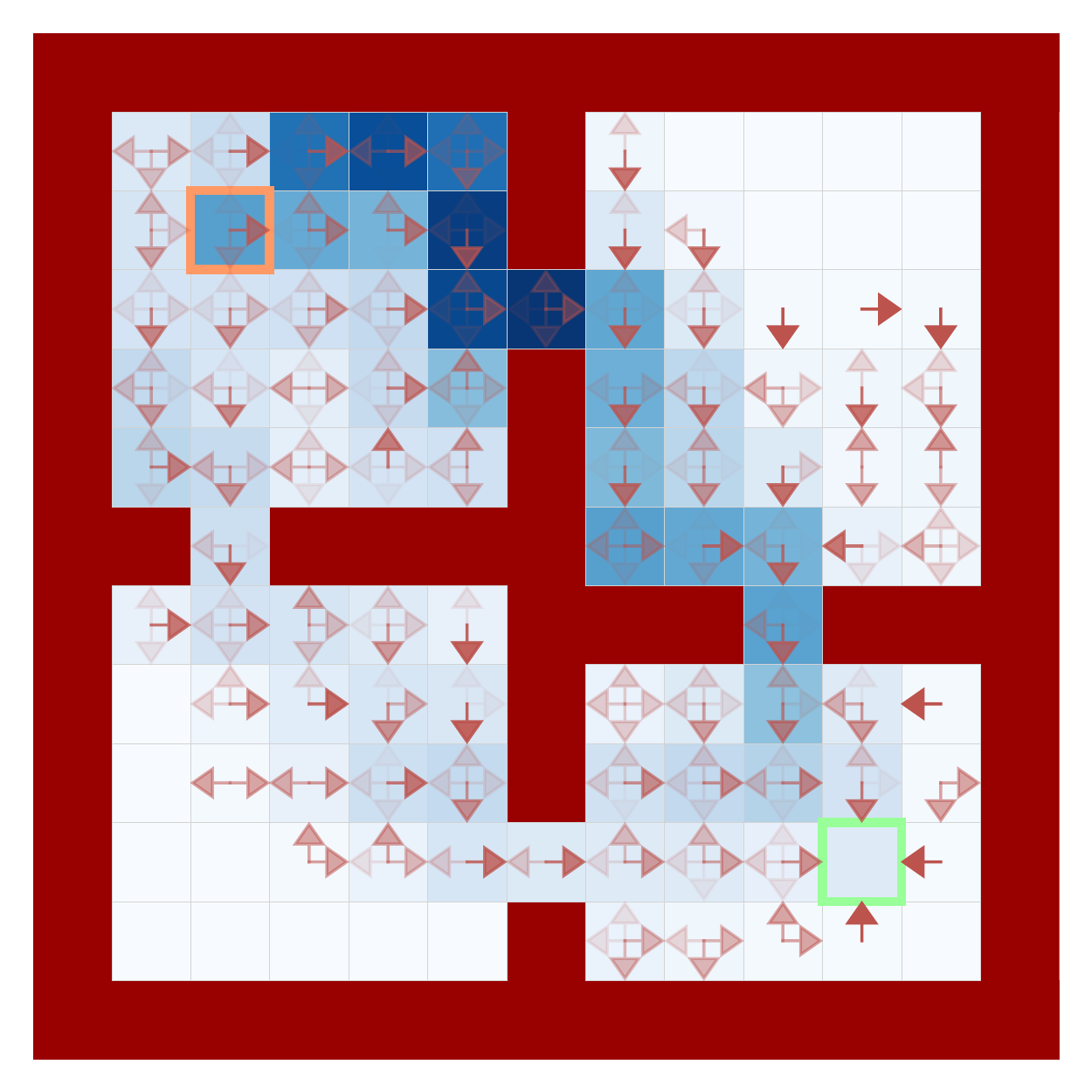}}
\subfigure[OptiDICE $w^*$]{\label{figure:illustrative_example:c}\includegraphics[width=0.49\columnwidth]{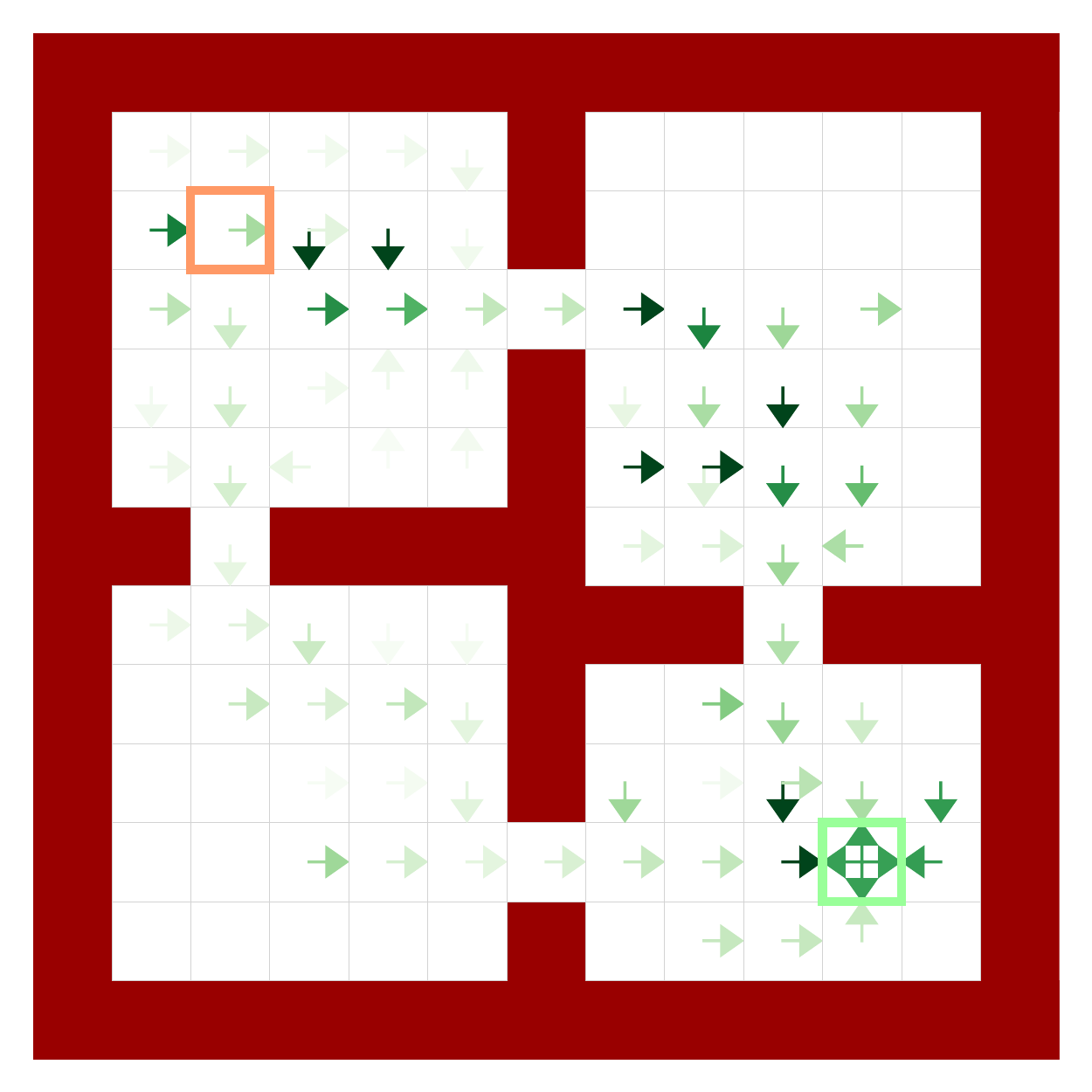}}
\subfigure[Estimated $\hat d^{\pi^*}$]{\label{figure:illustrative_example:d}\includegraphics[width=0.49\columnwidth]{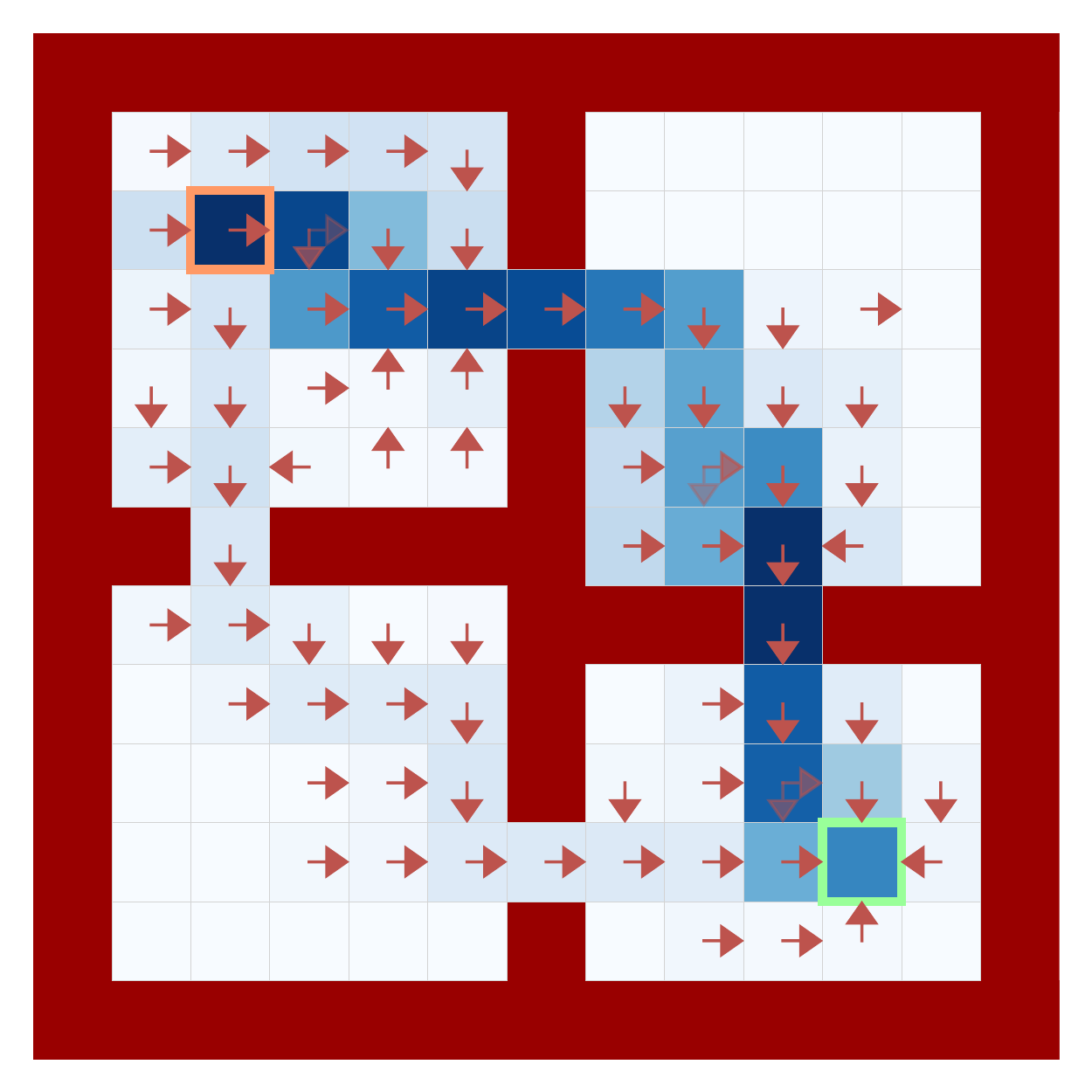}}
\vspace{-0.1in}
\caption{Illustrative example of how OptiDICE estimates the optimal policy's stationary distribution in the Four Rooms domain \citep{sutton1999between,nachum2019algaedice}. The initial state and the goal state are denoted by orange and green squares, respectively. Based on a sub-optimal data-collection policy $\piD$, which induces $d^{\piD}$ shown in (a), a static dataset is sampled and its empirical distribution $d^D$ ($\ne d^{\piD}$) shown in (b). The opacity of each square is determined by the state marginals of each stationary distribution, where the opacity of the arrow shows the policy induced by each stationary distribution. By multiplying the OptiDICE $w^*$ obtained by solving \eqref{eq:L_maxmin} (shown in (c)), a near-optimal policy $\pi^*$ is obtained from 
 $\hat d^{\pi^*}(s,a) = d^D(s,a) w^*(s,a)$ shown in (d).}
\vspace{-0.10in}
\label{figure:illustrative_example}
\end{figure*}

\subsection{A closed-form solution}
\label{subsection:closed_form_solution}
When the state and/or action spaces are large or continuous,
it is a standard practice to use function
approximators to represent terms such as $w$ and $\nu$,
and perform gradient-based optimization of $L$.
However, this could break nice properties for
optimizing $L$,
such as concavity in $w$ and convexity in $\nu$,
which causes numerical instability and poor convergence
for the maximin optimization~\cite{goodfellow2014generative}. 
We mitigate this issue by obtaining the closed-form
solution of the inner optimization, 
which reduces the overall problem into a 
unconstrained convex optimization.


Since the optimization problem~(\ref{eq:objective_d}-\ref{eq:objective_d_constraint2}) is an instance of convex optimization, one can easily show that the strong duality holds by the Slater's condition~\cite{boyd2004convex}. Hence we can reorder the optimization from maximin to minimax:
\begin{align}
    \min_{\nu} \max_{w \ge 0} L(w, \nu).
    \label{eq:L_minmax}
\end{align}

Then, for any $\nu$, a closed-form solution to the inner maximization of~\eqref{eq:L_minmax} can be derived as follows:
\begin{proposition}\label{proposition:closed_form_solution_v0}
The closed-form solution of the inner maximization of~\eqref{eq:L_minmax}, $w_\nu^* := \argmax_{w \ge0} L(w, \nu)$, is
\begin{align}
    w_\nu^*(s, a) =
    \max \left( 0, (f')^{-1}\left(\frac{e_\nu(s, a)}{\alpha}\right) \right) ~~ \forall s, a,
    \label{eq:w_analytic_solution}
\end{align}
where $(f')^{-1}$ is the inverse function of the derivative $f'$ of $f$ and is strictly increasing by strict convexity of $f$. (Proof in Appendix~\ref{appendix:proof_of_closed_form_solution_v0}.)
\end{proposition}
A closer look at~\eqref{eq:w_analytic_solution} reveals that 
that for a fixed $\nu$, the optimal stationary distribution correction $w_{\nu}^*(s, a)$ is larger for
a state-action pair with larger 
advantage $e_\nu(s, a)$.
This solution property has a natural interpretation 
as follows.
As $\alpha \rightarrow 0$, the term in \eqref{eq:objective_lagrangian_d2} becomes the Lagrangian of the primal LP for solving the MDP, where $d(s,a)$ serve as Lagrange multipliers to impose 
constraints $R(s,a) + \gamma (\T \nu)(s,a) \le \nu(s)~\forall s, a$. 
Also, each $\nu(s)$ serves as the optimization variable representing the optimal state value function~\cite{puterman1994markov}.
Thus, $e_{\nu^*}(s,a) = Q^*(s,a) - V^*(s)$, i.e.
the advan-\allowbreak tage function of the \emph{optimal policy},
should be zero for the optimal action while 
it should be lower for sub-optimal actions.
For $\alpha > 0$, the convex regularizer $f\big(\frac{d(s,a)}{d^D(s,a)}\big)$ in \eqref{eq:objective_lagrangian_d2} relaxes those constraints into soft ones, 
but it still prefers the actions with higher $e_{\nu^*}(s,a)$ over those with lower $e_{\nu^*}(s,a)$.
From this perspective, $\alpha$ adjusts the softness of the constraints $e_\nu(s,a)\le0~\forall s, a$, and $f$ determines the relation between advantages and 
stationary distribution corrections. 

Finally, we reduce the nested optimization in~\eqref{eq:L_minmax} to the following single minimization problem by plugging $w_\nu^*$ into $L(w,\nu)$:
\begin{align}
    &\min_{\nu} L(w_\nu^*, \nu)
    = (1 - \gamma) \E_{s \sim p_0}[\nu(s)] \label{eq:objective_nu} \\
    &~~~+ \E_{(s,a) \sim d^D} \left[ - \alpha f \Big( \max \Big(0,  (f')^{-1} \big( \tfrac{1}{\alpha} e_\nu(s,a) \big) \Big) \Big) \right] \nonumber \\
    &~~~+ \E_{(s,a) \sim d^D} \left[ \max \Big(0,  (f')^{-1}\big( \tfrac{1}{\alpha} e_\nu(s,a) \big) \Big) \big( e_\nu(s,a) \big) \right]. \nonumber
\end{align}

\begin{proposition}\label{proposition:L_v_convex}
    $L(w_{\nu}^*, \nu)$ is convex with respect to $\nu$. 
    (Proof in  Appendix~\ref{appendix:proof_of_L_v_convex}.)
\end{proposition}
The minimization of this convex objective can be performed much more reliably than the nested minimax optimization problem.
For practical purposes, we use the following
objective that can be easily optimized via sampling from $D$:
\begin{align}
    &\tilde L(\nu) := (1 - \gamma) \E_{s \sim p_0}[\nu(s)] \label{eq:objective_nu_biased} \\
    &+\E_{(s,a,s') \sim d^D}\bigg[-  \alpha f \Big( \max \Big(0,  (f')^{-1} \big( \tfrac{1}{\alpha} \hat e_\nu(s,a,s') \big) \Big) \Big) \nonumber \\
    &+ \max \Big(0,  (f')^{-1}\big( \tfrac{1}{\alpha} \hat e_\nu(s,a,s') \big) \Big) \big( \hat e_\nu(s,a,s') \big) \bigg]. \nonumber
\end{align}
However, careful readers may notice that 
$\tilde L(\nu)$ can be a biased estimate of our target objective $L(w_{\nu}^*, \nu)$ in \eqref{eq:objective_nu} due to non-linearity of $(f')^{-1}$ and double-sample problem \cite{Baird1995residual} in $L(w_\nu^*, \nu)$.
We justify $\tilde L(\nu)$ by formally showing that $\tilde L(\nu)$ is the upper bound of $L(w_{\nu}^*, \nu)$:
\begin{corollary}
\label{corollary:upper_bound}
$\tilde L(\nu)$ in \eqref{eq:objective_nu_biased} is an upper bound of $L(w_\nu^*, \nu)$ in \eqref{eq:objective_nu}, i.e. $L(w_\nu^*, \nu) \le \tilde L(\nu)$ always holds, where equality holds when the MDP is deterministic. (Proof in Appendix~\ref{appendix:proof_of_L_v_convex}.)
\end{corollary}

\vspace{-0.18in}
\paragraph{Illustrative example}

\figureref{figure:illustrative_example} outlines how our approach works in the Four Rooms domain \citep{sutton1999between} where the agent aims to navigate to a goal location in a maze composed of four rooms. We collected static dataset $D$ consisting of $50$ episodes with maximum time step $50$ using the data-collection policy $\piD=0.5\pi_\mathrm{true}^*+0.5\pi_{\mathrm{rand}}$, where $\pi_\mathrm{true}^*$ is the optimal policy of the underlying true MDP and $\pi_{\mathrm{rand}}$ is the random policy sampled from the Dirichlet distribution, i.e. $\pi_{\mathrm{rand}}(s)\sim \mathrm{Dir}(1,1,1,1)~\forall s$.

In this example, we explicitly constructed Maximum Likelihood Estimate (MLE) MDP $\hat M$ based on the static dataset $D$.
We then obtained $\nu^*$ by minimizing ~\eqref{eq:objective_nu} 
where $\hat M$ was used to exactly compute
$e_\nu(s,a)$.
Then, $w_{\nu^*}^*$ was estimated directly via Eq.~\eqref{eq:w_analytic_solution} (\figureref{figure:illustrative_example:c}).
Finally, $w^*$ was multiplied by $d^D$ to correct $d^D$ towards an optimal policy, resulting in $\hat d^{\pi^*}$, which is the stationary distribution of the estimated optimal policy (\figureref{figure:illustrative_example:d}).
For tabular MDPs, the global optima $(\nu^*, w^*)$ can always be obtained.
We describe these experiments on OptiDICE for finite MDPs in Appendix \ref{appendix:optidice_for_finite_mdp}.

\vspace{-0.12in}
\subsection{Stationary distribution correction estimation with function approximation}

Based on the results from the previous section, 
we assume that $\nu$ and $w$ are parameterized by $\theta$ and $\phi$, respectively, and that both models are sufficiently expressive, e.g. using deep neural networks.
Using these models, we optimize $\theta$ by
\begin{align}
    &\min_\theta J_\nu(\theta) := \min_\theta \tilde L (\nu_\theta). \label{eq:objective_nu_theta}
\end{align}
After obtaining the optimizing solution $\theta^*$, 
we need a way to evaluate $w^*(s,a) = \frac{d^{\pi^*}(s,a)}{d^D(s,a)}$
for any $(s,a)$ to finally obtain the optimal policy $\pi^*$.
However, the closed-form solution $w_{\nu_{\theta}}^*(s,a)$ in 
\propositionref{proposition:closed_form_solution_v0} can be evaluated
only on $(s,a)$ in $D$ since it requires both $R(s,a)$ and $\E_{s' \sim T(s,a)}[ \nu_{\theta}(s') ]$ to evaluate the advantage $e_\nu$. 
Therefore, 
we use a parametric model $e_\phi$ that 
approximates the advantage inside the
analytic formula presented in~\eqref{eq:w_analytic_solution}, so that 
\begin{align}
    w_\phi(s, a) := \max \left( 0, 
        (f')^{-1}\left(\frac{e_\phi(s, a)}{\alpha}\right)
    \right).
    \label{eq:estimator_e}
\end{align}
We consider two options to optimize $\phi$ once we
obtain $\nu_\theta$ from Eq.~\eqref{eq:objective_nu_theta}.
First, $\phi$ can be optimized via
\begin{align}
    \min_{\phi} J_{w}(\phi ; \theta) := \min_{\phi} -\hat L (w_\phi, \nu_{\theta}) \label{eq:objective_e_minimax}
\end{align}
which corresponds to solving the original minimax problem \eqref{eq:L_minmax}. We also consider
\begin{align}
    \min_{\phi}&~J_w^\mathrm{MSE}(\phi ; \theta) \nonumber \\
    &\hspace{-3ex}:= \min_\phi \E_{\substack{(s, a, s') \sim d^D}} \Big[  \big( e_\phi(s, a) - \hat e_{\nu_{\theta}}(s, a, s') \big)^2 \Big],
    \label{eq:objective_e_mse}
\end{align}
which minimizes the mean squared error (MSE) between the advantage $e_\phi(s,a)$ and the target induced by $\nu_\theta$.
We observe that using either $J_w$ or $J_w^\mathrm{MSE}$ works effectively, which will be detailed in our experiments.
In our implementation, we perform joint training of $\theta$ and $\phi$, rather than optimizing $\phi$ after convergence of $\theta$.

\subsection{Policy extraction}
\label{subsection:policy_extraction}
As the last step,
we need to extract the optimal policy $\pi^*$ from the optimal stationary distribution corrections $w_\phi(s,a) = \frac{d^{\pi^*}(s,a)}{d^D(s,a)}$.
While the optimal policy can be easily obtained by $\pi^*(a|s) = \frac{ d^D(s,a) w_\phi(s,a) }{\sum_{\bar a} d^D(s,\bar a) w_\phi(s, \bar a)}$ for tabular domains, this procedure is not straightforwardly applicable to continuous domains.

One of the ways to address continuous domains is to
use importance-weighted behavioral cloning:
we optimize the parameterized policy $\pi_\psi$ by maximizing the log-likelihood on $(s,a)$ 
that would be sampled from the optimal policy $\pi^*$:
\begin{align*}
    \max_\psi
    &~\E_{(s, a)\sim d^{\pi^*}}[\log\pi_\psi(a|s)]\\
    &~~~~~~~=
    \max_\psi \E_{(s, a)\sim d^D}
    \left[
        \textstyle w_\phi(s, a)\log\pi_\psi(a|s)
    \right].
\end{align*}
Despite its simplicity, this approach does not 
work well in practice, since  
$\pi_\psi$ will be trained only on samples from the
intersection of the supports of $d^{\pi^*}$ 
and $d^D$, which becomes very scarce
when $\pi^*$ deviates significantly 
from the data collection policy $\piD$.


We thus use the information projection (I-projection)
for training the policy:
\begin{align}
    \min_{\psi}~
	\mathbb{KL}
	\left(
	    d^D(s) \pi_\psi(a|s) || d^D(s) \pi^*(a|s)
	\right),
	\label{eq:iprojection_kl}
\end{align}
where we replace $d^{\pi^*}(s)$ by $d^D(s)$ for $d^{\pi^*}(s, a)$.
This results in minimizing the discrepancy between $\pi_\psi(a|s)$ and $\pi^*(a|s)$ on the stationary
distribution over states from $\piD$.
This approach is motivated by the desideratum that
the policy $\pi_\psi$ should be trained at 
least on the states observed in $D$ to be robust upon deployment.
Now, rearranging the terms in~\eqref{eq:iprojection_kl}, we obtain
\begin{align}
	&
	\mathbb{KL}\left( d^D(s) \pi_\psi(a|s) || d^D(s) \pi^*(a|s) \right)
	\nonumber
	\\
	&\hspace{-0.04in}=
	-\E_{\hspace{-1.6pt}\substack{s\sim d^D\\a\sim\pi_\psi(s)}}
	\bigg[
	    \log
	    \underbrace{\frac{d^*(s, a)}{d^D(s, a)}}_{=w_\phi(s, a)}
	    -
	    \log\frac{\pi_\psi(a|s)}{\piD(a|s)}
	    -
	    \underbrace{\log\frac{d^*(s)}{d^D(s)}}_{\text{constant for $\pi$}}
	\bigg]
	\nonumber
	\\
	&\hspace{-0.04in}
	=
	-\E_{\hspace{-1.6pt}\substack{s\sim d^D\\a\sim\pi_\psi(s)}}
	\left[ 
	    \log w_\phi(s, a)
	    -
	    \KL(\pi_\psi(\bar{a}|s)||\piD(\bar{a}|s))
	\right]
	+C
	\nonumber
	\\
	&\hspace{-0.04in}
	=:
	J_\pi(\psi;\phi, \piD)
	\label{eq:objective_psi}
\end{align}
We can interpret this I-projection objective 
as a KL-regularized actor-critic architecture~\citep{fox2016taming,schulman2017equivalence}, where
$\log w_\phi(s,a)$ taking the role of the critic 
and $\pi_\psi$ being the actor\footnote{
When $f(x) = x \log x$ (i.e. KL-divergence), $(f')^{-1} = \exp(x - 1)$, and we have $\log w_{\nu^*}(s,a) = \frac{1}{\alpha} e_{\nu^*}(s,a) - 1$ by Eq.~\eqref{eq:w_analytic_solution}.
Given that $e_{\nu^*}(s,a)$ represents an approximately optimal advantage $A^*(s,a) \approx Q^*(s,a) - V^*(s)$ (Section \ref{subsection:closed_form_solution}), the policy extraction via I-projection \eqref{eq:objective_psi} corresponds to a KL-regularized policy optimization: $\max_\pi \E_{a \sim \pi}[ \frac{1}{\alpha} A^*(s,a) - KL(\pi(\bar a | s) || \pi_D(\bar a | s)) ]$.}.
Note that I-projection requires us to evaluate $\piD$ for the KL regularization term.
For this, we employ another parameterized policy
$\pi_\beta$ to approximate $\piD$, trained via
simple behavioral cloning (BC). 

\subsection{Generalization to $\gamma=1$}
\label{subsection:generalization_to_gamma1}

For $\gamma=1$, our original problem (\ref{eq:objective_d}-\ref{eq:objective_d_constraint2}) for the stationary distribution $d$ is an ill-posed problem: for any $d$ that satisfies the Bellman flow constraints (\ref{eq:objective_d_constraint1}-\ref{eq:objective_d_constraint2}) and a constant $c\ge0$, $cd$ also satisfies the Bellman flow constraints (\ref{eq:objective_d_constraint1}-\ref{eq:objective_d_constraint2})~\cite{zhang2019gendice}. 
We address this issue by adding additional normalization constraint $\sum_{s,a}d(s, a)=1$ to (\ref{eq:objective_d}-\ref{eq:objective_d_constraint2}).
By using analogous derivation from \eqref{eq:objective_d} to \eqref{eq:L_minmax} with the normalization constraint---introducing a Lagrange multiplier $\lambda\in\R$ and changing the variable $d$ to $w$---we obtain the following minimax objective for $w$, $\nu$ and $\lambda$:
\begin{align}
    &\min_{\nu,\lambda}
    \max_{w\ge0}
    L(w,\nu,\lambda)\nonumber
    \\
    &\hspace{10pt}
    :=
    L(w, \nu)+\lambda(1-\E_{(s, a)\sim d^D}[w(s, a)])
    \nonumber\\
    &\hspace{10pt}
    =
    (1-\gamma)\E_{s\sim p_0}[\nu(s)]
    +
    \E_{(s, a)\sim d^D}[-\alpha f(w(s, a))]
    \nonumber\\
    &\hspace{10pt}
    ~~~~~
    -\E_{(s, a)\sim d^D}[w(s, a)(e_\nu(s, a)-\lambda))]
    +\lambda.
    \label{eq:mcL_minmax}
\end{align}

Similar to \eqref{eq:objective_minmax_sars1}, we define $\hat{L}(w, \nu, \lambda)$, an unbiased estimator for $L(w, \nu, \lambda)$ such that
\begin{align}
    &\hat L(w, \nu, \lambda) := (1 - \gamma) \E_{s \sim p_0}[\nu(s)] 
    +\lambda
    \label{eq:objective_minmax_sars1_lambda} \\
    &+ \E_{(s,a,s') \sim d^D\hspace{-1.7pt}} \left[- \alpha f \big( w(s,a) \big) + w(s,a) \big( \hat e_{\nu,\lambda}(s,a,s') \big) \right]\nonumber,
\end{align}
where $\hat{e}_{\nu,\lambda}(s, a, s') := \hat{e}_{\nu}(s, a, s')-\lambda$.
We then derive a closed-form solution for the inner maximization in \eqref{eq:mcL_minmax}:
\begin{proposition}\label{proposition:closed_form_solution_v1}
The maximizer $w_{\nu,\lambda}^*:\SA\rightarrow\R$ of the inner optimization of \eqref{eq:mcL_minmax}, which is defined by $w_{\nu,\lambda}^*:=\argmax_{w\ge0}L(w, \nu, \lambda)$, is 
\begin{align*}
    w_{\nu,\lambda}^*(s, a)
    =
    \max
    \left(
        0, (f')^{-1}\left(\frac{e_\nu(s, a)-\lambda}{\alpha}\right)
    \right).
\end{align*}
(Proof in Appendix~\ref{appendix:proof_of_closed_form_solution_v1}.)
\end{proposition}

Similar to \eqref{eq:objective_nu_biased}, we minimize 
the biased estimate $\tilde{L}(\nu,\lambda)$, which is an upper bound of $L(w_{\nu,\lambda}^*,\nu,\lambda)$,
by applying the closed-form solution from \propositionref{proposition:closed_form_solution_v1}:
\begin{align}
    &\tilde L(\nu, \lambda) := (1 - \gamma) \E_{s \sim p_0}[\nu(s)]  \label{eq:objective_nu_lambda_biased} \\
    &\hspace{-0.04in}+ \E_{(s,a,s') \sim d^D} \bigg[\hspace{-2pt}- \alpha f \Big( \max \big(0,  (f')^{-1} \big( \tfrac{1}{\alpha} \hat{e}_{\nu,\lambda}(s, a, s') \big)\big) \Big) \nonumber \\
    &\hspace{-0.04in}+ \max \Big(0,  (f')^{-1}\big( \tfrac{1}{\alpha} \hat{e}_{\nu,\lambda}(s, a, s') \big) \Big) \big( \hat{e}_{\nu,\lambda}(s, a, s') \big) \bigg]+\lambda.\nonumber
\end{align}
By using the above estimators, we correspondingly update our previous objectives for $\theta$ and $\phi$ as follows.
First, the objective for $\theta$ is modified to 
\begin{align}
    \min_{\theta}
    J_\nu(\theta, \lambda)
    :=
    \min_{\theta}
    \tilde{L}(\nu_\theta,\lambda).
    \label{eq:objective_theta}
\end{align}
For $\phi$, we modify our approximator in \eqref{eq:estimator_e} by including the Lagrangian $\lambda'\in\R$:
\begin{align*}
    w_{\phi,\lambda'}(s, a)
    :=
    \max
    \left(
        0, 
        (f')^{-1}\left(\frac{e_\phi(s, a)-\lambda'}{\alpha}\right)
    \right).
\end{align*}
Note that $\lambda'\ne\lambda$ is used to stabilize the learning process. 
For optimizing over $\phi$, the minimax objective \eqref{eq:objective_e_minimax} is modified as
\begin{align}
    \min_\phi
    J_w(\phi, \lambda';\theta)
    :=
    \min_\phi
    -
    \hat{L}(w_{\phi,\lambda'}, \nu_\theta, \lambda'),
    \label{eq:objective_e_minimax_lambda_prime}
\end{align}
while the same objective $J_w^{\mathrm{MSE}}(\phi;\theta)$ in \eqref{eq:objective_e_mse} is used for the MSE objective. 
We additionally introduce learning objectives for $\lambda$ and $\lambda'$, which is required for the normalization constraint discussed in this subsection:
\begin{align}
    \min_{\lambda}
    J_\nu(\theta,\lambda)
    ~~~\text{ and }~~~
    \min_{\lambda'}
    J_w(\phi, \lambda';\theta).
    \label{eq:objective_lambda}
\end{align}
Finally, by using the above objectives in addition to BC objective and policy extraction objective in \eqref{eq:objective_psi}, we describe our algorithm, OptiDICE, in \textbf{Algorithm~\ref{alg:optidice}}, where we train neural network parameters via stochastic gradient descent.
In our algorithm, we use a warm-up iteration---optimizing all networks except $\pi_\psi$---to prevent $\pi_\psi$ from its converging to sub-optimal policies during its initial training. 
In addition, we empirically observed that using the normalization constraint stabilizes OptiDICE's learning process even for $\gamma<1$, thus we
used the normalization constraint in all experiments~\cite{zhang2019gendice}.

\begin{algorithm}[t!]
\caption{OptiDICE}
\centering
\begin{algorithmic}[1]
	\REQUIRE 
	A dataset $D:=\{(s_i, a_i, r_i, s_i')\}_{i=1}^N$, 
	a set of initial states $D_0:=\{s_{0,i}\}_{i=1}^{N_0}$,
	neural networks $\nu_\theta$ and $e_\phi$ with parameters $\theta$ and $\phi$,
	learnable parameters $\lambda$ and $\lambda'$,
	policy networks $\pi_\beta$ and $\pi_\psi$ with parameter $\beta$ and $\psi$, a learning rate $\eta$
	\FOR{each iteration}
	    \STATE
	    Sample mini-batches from $D$ and $D_0$, respectively.
	    
	    \STATE
	    Compute $\theta$-gradient to optimize \eqref{eq:objective_theta}:
	    \vspace{-0.1in}
	    \begin{align*}
	        g_\theta
	        \approx
	        \nabla_\theta 
	        J_\nu(\theta, \lambda)
	    \end{align*}
	    \vspace{-0.25in}
	    
	    \STATE
	    Compute $\phi$-gradient for either one of objectives:
	    \vspace{-0.1in}
	    \begin{align*}
	        g_\phi
	        &\approx
	        \nabla_\phi 
	        J_w(\phi, \lambda';\theta)
	        &&\text{(\emph{minimax obj. \eqref{eq:objective_e_minimax_lambda_prime}})}\\
	        g_\phi
	        &\approx
	        \nabla_\phi J_{w}^{\mathrm{MSE}}(\phi ; \theta)
	        &&\text{(\emph{MSE obj. \eqref{eq:objective_e_mse}})}
	    \end{align*}
	    \vspace{-0.25in}
	    
	    \STATE
	    Compute $\lambda$ and $\lambda'$ gradients to optimize \eqref{eq:objective_lambda}:
	    \vspace{-0.1in}
	    \begin{align*}
	        g_\lambda
	        \approx 
	        \nabla_\lambda 
	        J_\nu(\theta, \lambda), 
	        g_{\lambda'}
	        \approx 
	        \nabla_{\lambda'}
	     J_w(\phi,\lambda';\theta)
	    \end{align*}
	    \vspace{-0.25in}
	    
	    \STATE
	    Compute $\beta$-gradient $g_\beta$ for BC.
	    
        \STATE
        Compute $\psi$-gradient via \eqref{eq:objective_psi} (\emph{policy extraction}):
        \vspace{-0.1in}
        \begin{align*}
	        g_\psi
	        \approx 
	        \nabla_\psi J_\pi(\psi;\phi,\pi_\beta)
        \end{align*}
	    \vspace{-0.25in}
	    
	    \STATE
	    Perform SGD updates:\\
        \vspace{-0.15in}~~
        \begin{minipage}{.25\columnwidth}
            \begin{align*}
                \theta
                &\leftarrow\theta-\eta g_\theta,\\
                \phi
                &\leftarrow\phi-\eta g_\phi,
            \end{align*}
        \end{minipage}
        ~
        \begin{minipage}{.28\columnwidth}
            \begin{align*}
                \lambda
                &\leftarrow\lambda-\eta g_\lambda,\\
                \lambda'
                &\leftarrow\lambda'-\eta g_{\lambda'},
            \end{align*}
        \end{minipage}
        ~
        \begin{minipage}{.26\columnwidth}
            \begin{align*}
                \beta
                &\leftarrow\beta-\eta g_\beta,\\
                \psi
                &\leftarrow\psi-\eta g_\psi.
            \end{align*}
        \end{minipage}
	\ENDFOR
	\ENSURE $\nu_\theta\approx \nu^*$, $w_{\phi, \lambda'}\approx w^*$, $\pi_\psi\approx\pi^*$, 
\end{algorithmic}
\label{alg:optidice}
\end{algorithm}

\begin{figure*}[t!]
    \centering
    \includegraphics[width=1.0\textwidth]{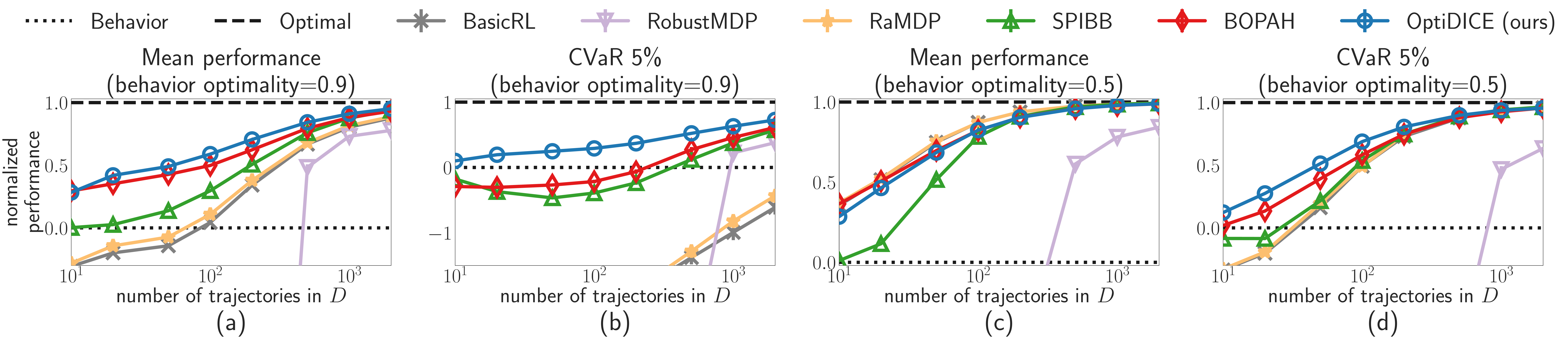}
	\vspace{-0.2in}
    \caption{Performance of tabular OptiDICE and baseline algorithms in random MDPs. 
    For baselines, we use
    \emph{BasicRL} (a model-based RL computing an optimal policy via MLE MDP), \emph{Robust MDP} \cite{Nilim2005robust,Iyengar2005robust}, \emph{Reward-adjusted MDP (RaMDP)} \cite{petrik2016safe}, \emph{SPIBB} \cite{laroche2019safe}, \emph{BOPAH} \cite{lee2020batch}. 
    For varying numbers of trajectories and two types of data-collection policies ($\zeta=0.9, 0.5$), the mean and the 5\%-CVaR of normalized performances for 10,000 runs are reported with 95\% confidence intervals.
    OptiDICE performs better than (for $\zeta=0.9$) or on par with (for $\zeta=0.5$) the baselines in the mean performance measure, while always outperforming the baselines in the CVaR performance measure.
    }
    \label{fig:random_mdp}
\end{figure*}

\section{Experiments}

In this section, we evaluate OptiDICE for both tabular and continuous MDPs.
For the $f$-divergence, we chose $f(x) = \frac{1}{2}(x-1)^2$, i.e. $\chi^2$-divergence for the tabular-MDP experiment,
while we use its softened version 
for continuous MDPs (See Appendix~\ref{appendix:f_divergence} for details).

\subsection{Random MDPs (tabular MDPs)}

We validate tabular OptiDICE's efficiency and robustness using randomly generated MDPs by following the experimental protocol from  \citet{laroche2019safe} and \citet{lee2020batch} (See Appendix~\ref{appendix:random_mdps}.).
We consider a data-collection policy $\piD$ characterized by the behavior optimality parameter $\zeta$ that relates to $\piD$'s performance $\zeta V^{*}(s_0) + (1 - \zeta) V^{\pi_\mathrm{unif}}(s_0)$
where $\pi_\mathrm{unif}$ denotes the uniformly
random policy.
We evaluate each algorithm in terms of the normalized performance of the policy $\pi$, given by $(V^{*}(s_0) - V^{\piD}(s_0))/(V^\pi(s_0) - V^{\piD}(s_0))$, which intuitively measures the performance enhancement of $\pi$ over $\piD$.
Each algorithm is tested for 10,000 runs, and their mean and 5\% conditional value at risk (5\%-CVaR) are reported, where the mean of the worst 500 runs is considered for 5\%-CVaR. Note that CVaR implicitly stands for the robustness of each algorithm.

We describe the performance of tabular OptiDICE and baselines in \figureref{fig:random_mdp}.
For $\zeta=0.9$, where $\piD$ is near-deterministic and thus $d^D$'s support is relatively small, OptiDICE outperforms the baselines in both mean and CVaR (\emph{Figure~\ref{fig:random_mdp}(a),(b)}).  
For $\zeta = 0.5$, where $\piD$ is highly stochastic and thus $d^D$'s support is relatively large, OptiDICE outperforms the baselines in CVaR, while performing competitively in mean. 
In summary, OptiDICE was more sample-efficient and stable than the baselines. 


\begin{table}[t!]
\caption{Normalized performance of OptiDICE compared with the best model-free baseline in the D4RL benchmark tasks~\citep{fu2020d4rl}. In the \emph{Best baseline} column, the algorithm with the best performance among 8 algorithms (offline SAC~\citep{haarnoja2018soft}, BEAR~\citep{kumar2019stabilizing}, BRAC~\citep{wu2019behavior}, AWR~\citep{peng2019advantage}, cREM~\citep{agarwal2020optimistic}, BCQ~\citep{fujimoto2019off}, AlgaeDICE~\cite{nachum2019algaedice}, CQL~\cite{kumar2020conservative}) is presented, taken from~\citep{fu2020d4rl}. OptiDICE achieved highest scores in 7 tasks.
}
\vspace{0.05in}
\label{table:optidice_and_baselines}
\begin{center}
\begin{small}
\begin{tabular}{l|r@{\hskip 0.02in}l@{\hskip -0.1in}rr}
\toprule
\multirow{2}{*}{D4RL Task} & \multicolumn{2}{c}{\multirow{2}{*}{\shortstack[c]{Best baseline}}}
& \multirow{2}{*}{OptiDICE} \\
& \multicolumn{2}{c}{} & \\
\midrule
\midrule
maze2d-umaze              & 88.2&\textsuperscript{Offline SAC}  & \textbf{111.0} \\
maze2d-medium             & 33.8&\textsuperscript{BRAC-v}       & \textbf{145.2} \\
maze2d-large              & 40.6&\textsuperscript{BRAC-v}       & \textbf{155.7} \\
\hline
hopper-random             & \textbf{12.2}&\textsuperscript{BRAC-v}& 11.2           \\
hopper-medium             & 58.0&\textsuperscript{CQL}  & \textbf{94.1}   \\
hopper-medium-replay      & \textbf{48.6}&\textsuperscript{CQL}  & 36.4    \\
hopper-medium-expert      & 110.9&\textsuperscript{BCQ} & \textbf{111.5}          \\
\hline
walker2d-random           & 7.3&\textsuperscript{BEAR}   & \textbf{9.9}   \\
walker2d-medium           & \textbf{81.1}&\textsuperscript{BRAC-v} & 21.8   \\
walker2d-medium-replay    & \textbf{26.7}&\textsuperscript{CQL} & 21.6   \\
walker2d-medium-expert    & \textbf{111.0}&\textsuperscript{CQL}& 74.8   \\
\hline
halfcheetah-random        & \textbf{35.4}&\textsuperscript{CQL} & 11.6 \\
halfcheetah-medium        & \textbf{46.3}&\textsuperscript{BRAC-v}  & 38.2  \\
halfcheetah-medium-replay & \textbf{47.7}&\textsuperscript{BRAC-v}  & 39.8 \\
halfcheetah-medium-expert & 64.7&\textsuperscript{BCQ}  & \textbf{91.1}   \\
\bottomrule
\end{tabular}
\end{small}
\end{center}
\vspace{-0.2in}
\end{table}

\begin{figure*}[t!]
    \centering
	\includegraphics[width=1.0\textwidth]{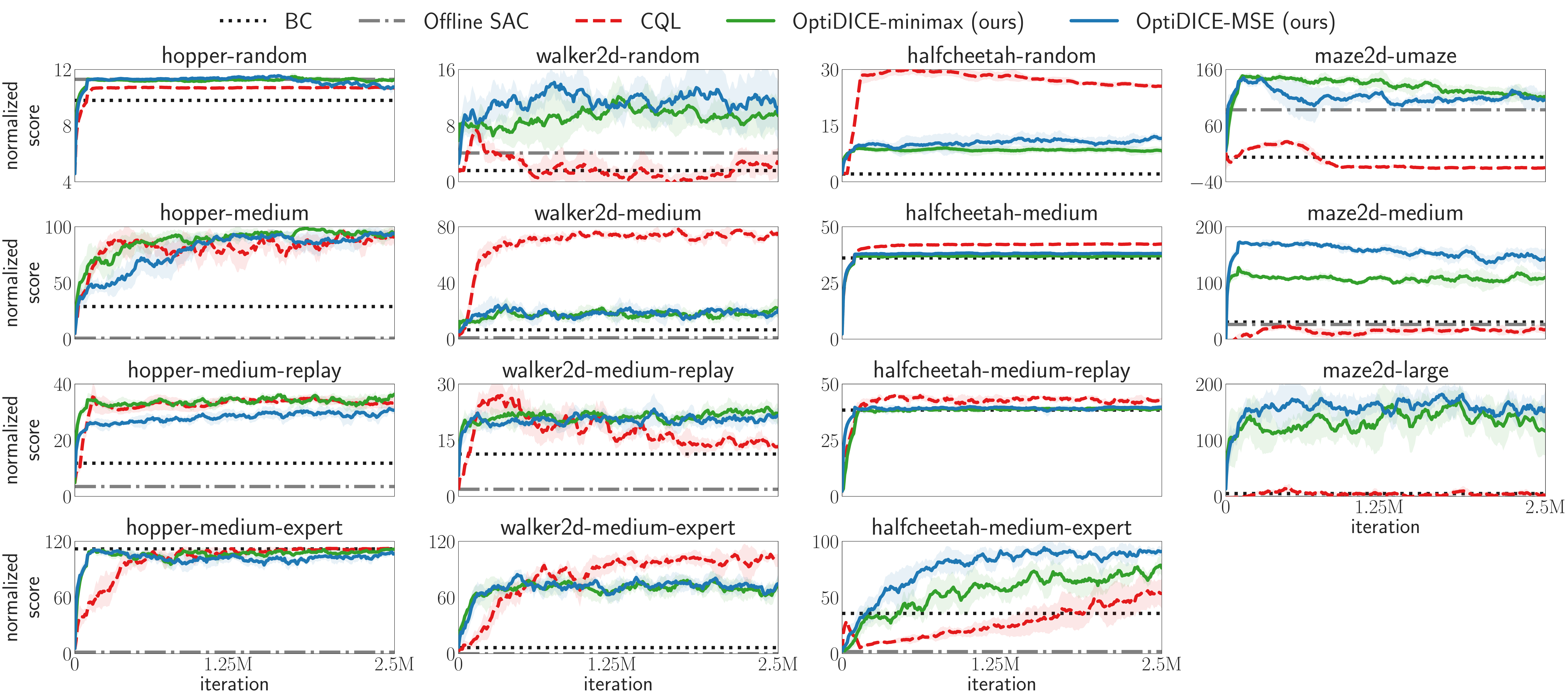}
	\vspace{-0.2in}
	\caption{Performance of BC, offline SAC~\citep{haarnoja2018soft}, CQL~\cite{kumar2020conservative}, 
	OptiDICE-minimax (= OptiDICE with minimax objective in~\eqref{eq:objective_e_minimax}) and OptiDICE-MSE (= OptiDICE with MSE objective  in~\eqref{eq:objective_e_mse}) on D4RL benchmark~\cite{fu2020d4rl} for $\gamma=0.99$. 
	For BC and offline SAC, we use the result reported in D4RL paper~\cite{fu2020d4rl}.
	For CQL and OptiDICE, we provide learning curves for each algorithm where the policy is optimized during 2,500,000 iterations. 
	For CQL, we use the original code by authors with hyperparameters reported in the CQL
	paper~\cite{kumar2020conservative}. 
	OptiDICE strictly outperforms CQL on 6 tasks, while performing on par with CQL on 4 tasks. We report mean scores and their 95\% confidence intervals obtained from 5 runs for each task.
	}
	\label{figure:d4rl_learning_curve}
\end{figure*}

\begin{figure*}[t!]
    \centering
	\includegraphics[width=1.0\textwidth]{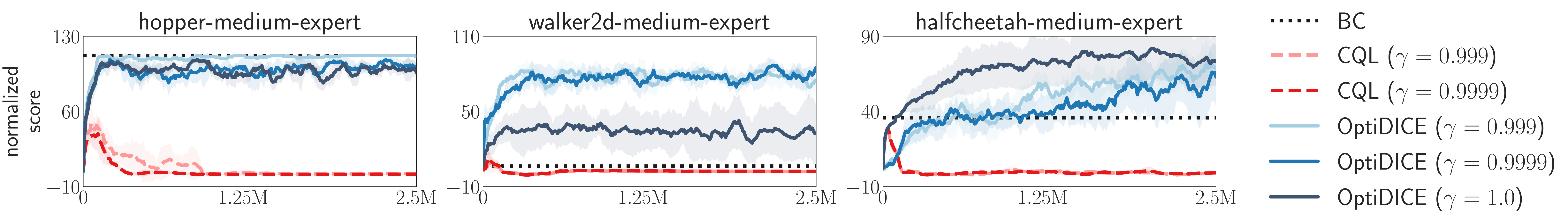}
	\vspace{-0.1in}
	\caption{
	Performance of BC, CQL and OptiDICE (with MSE objective in \eqref{eq:objective_e_mse}) in D4RL benchmark~\cite{fu2020d4rl} for $\gamma=0.999, 0.9999$ and $1.0$, where the hyperparameters other than $\gamma$ are the same as those in \figureref{figure:d4rl_learning_curve}.
	}
	\vspace{-0.2in}
	\label{figure:d4rl_learning_curve_gamma}
\end{figure*}

\subsection{D4RL benchmark (continuous control tasks)}

We evaluate OptiDICE in continuous MDPs using D4RL offline RL benchmarks~\cite{fu2020d4rl}.
We use Maze2D (3 tasks) and Gym-MuJoCo (12 tasks) domains from the D4RL dataset (See Appendix~\ref{appendix:d4rl_benchmark} for task description). 
We interpret terminal states as absorbing states and use the absorbing-state implementation proposed by~\citet{kostrikov2018discriminator}.
For obtaining $\pi_\beta$ discussed in Section 3.3, we use the tanh-squashed \emph{mixture} of Gaussians policy $\pi_\beta$ to embrace the multi-modality of data collected from heterogeneous policies.
For the target policy $\pi_\psi$, we use a tanh-squashed Gaussian policy, following conservative Q Learning (CQL)~\cite{kumar2020conservative}---the state-of-the-art model-free offline RL algorithm.
We provide detailed information of the experimental
setup in Appendix~\ref{appendix:d4rl_benchmark}.

The normalized performance of OptiDICE and the best model-free algorithm for each domain is presented in \emph{Table~\ref{table:optidice_and_baselines}}, and learning curves for CQL and OptiDICE are shown in \figureref{figure:d4rl_learning_curve}, where $\gamma = 0.99$ used for all algorithms.
Most notably, OptiDICE achieves state-of-the-art performance for all tasks in the Maze2D domain, by a large margin. 
In Gym-MuJoCo domain, OptiDICE achieves the best mean performance for 4 tasks (hopper-medium, hopper-medium-expert, walker2d-random, and halfcheetah-medium-expert).
Another noteworthy observation is that OptiDICE overwhelmingly outperforms AlgaeDICE~\cite{nachum2019algaedice} in all domains (\emph{Table~\ref{table:optidice_and_baselines}} and \emph{Table~\ref{table:appendix_optidice_and_baselines}} in Appendix for detailed performance of AlgaeDICE), although both AlgaeDICE and OptiDICE stem from the same objective in~\eqref{eq:objective_pi}.
This is because AlgaeDICE optimizes a nested max-min-max problem, which can suffer from severe overestimation by using out-of-distribution actions and numerical instability.
In contrast, OptiDICE solves a simpler minimization problem and does not rely on out-of-distribution actions, exhibiting stable optimization.

\begin{figure*}[t!]
    \centering
    \newcommand{\toyfigheight}{1.98cm}
    \includegraphics[height=\toyfigheight]{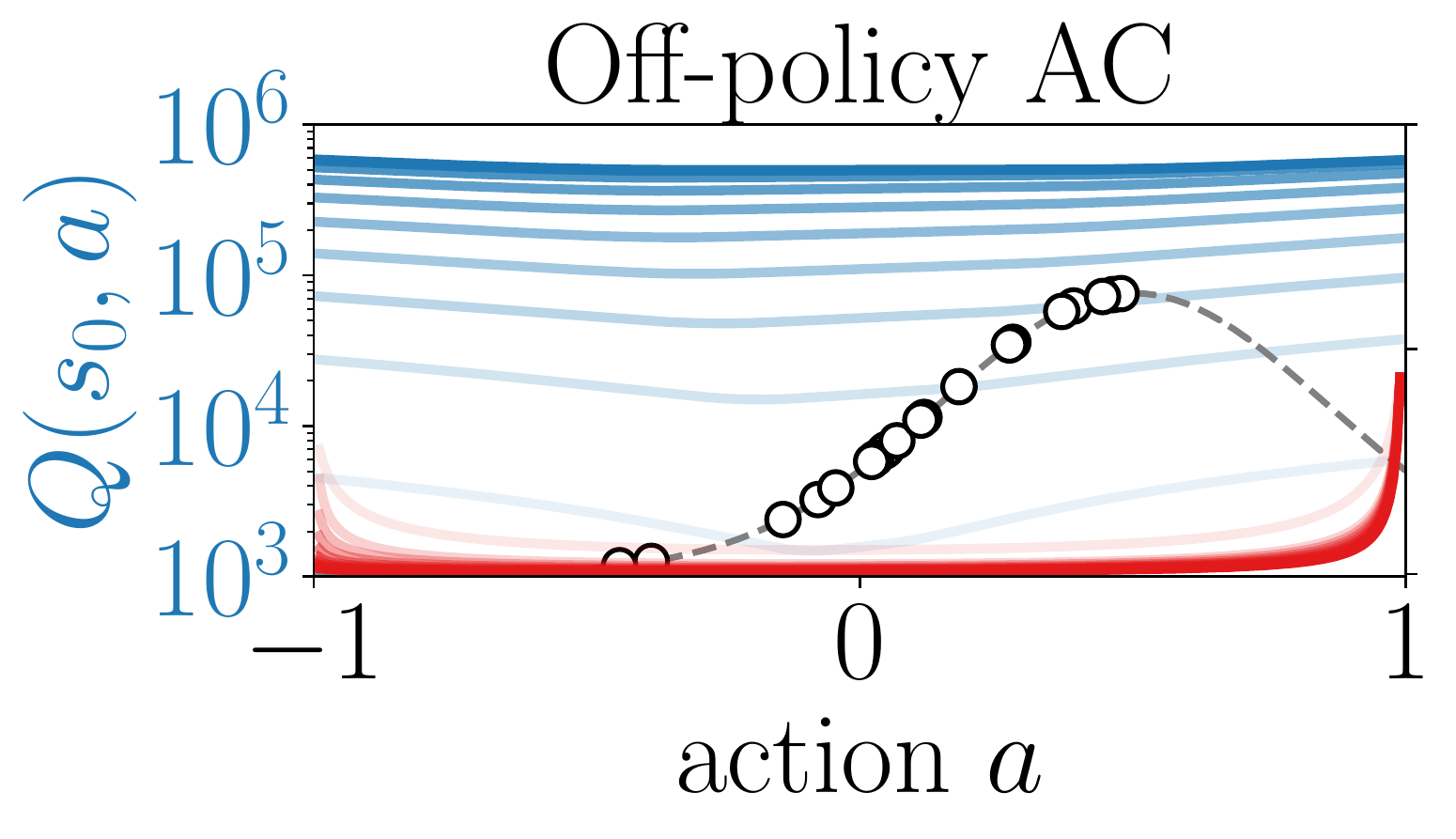}
    \includegraphics[height=\toyfigheight]{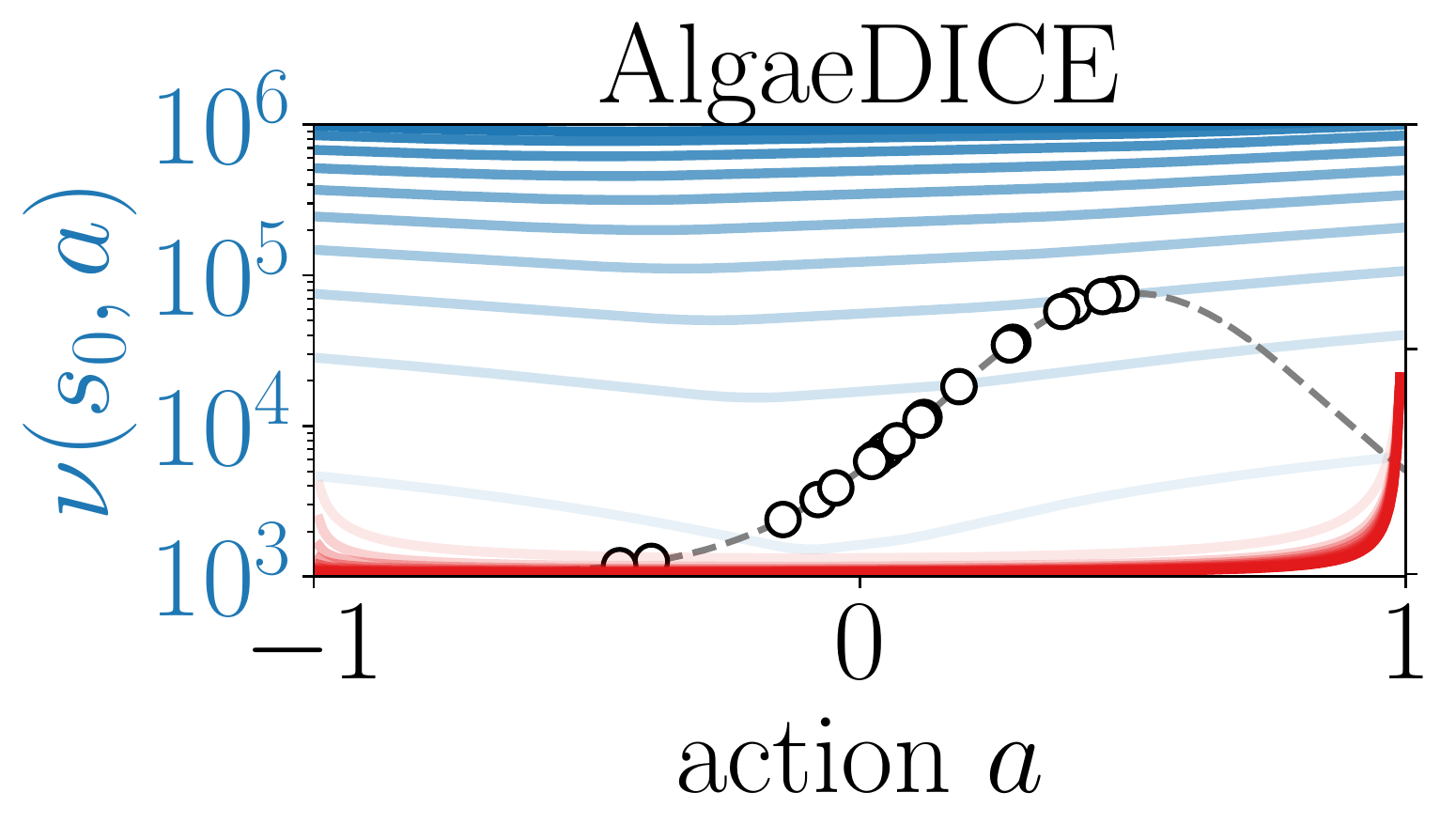}
    \includegraphics[height=\toyfigheight]{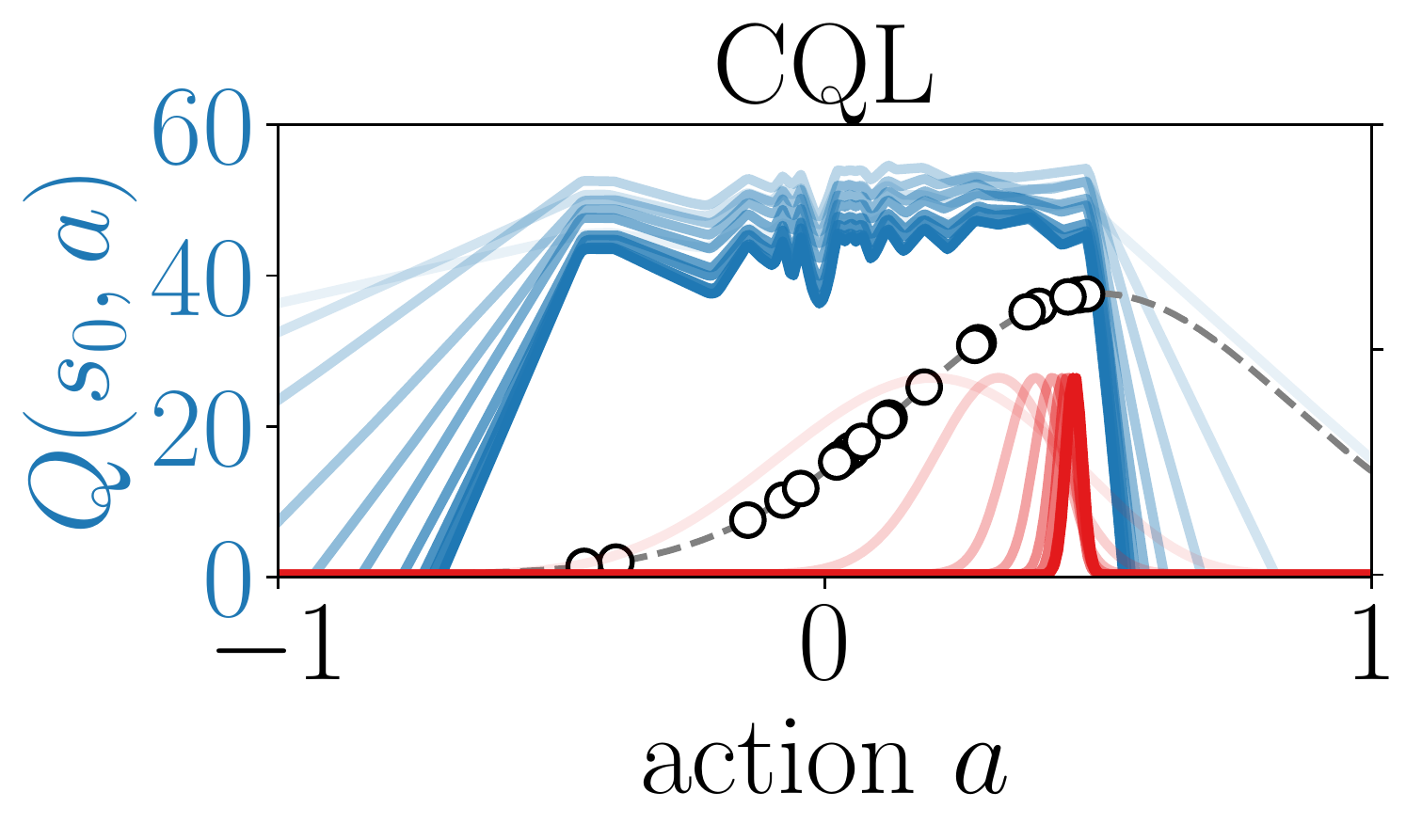}
    \includegraphics[height=\toyfigheight]{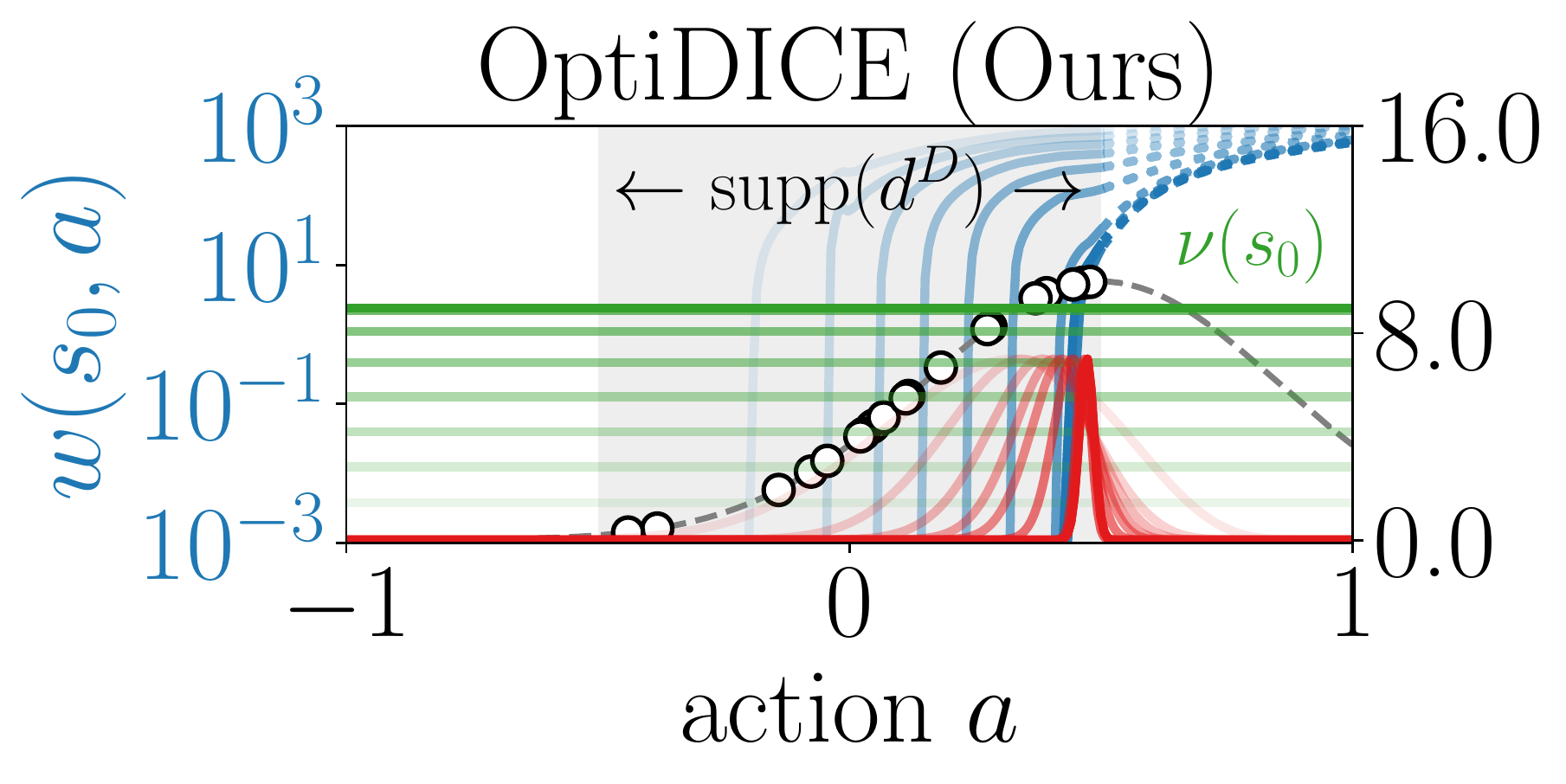}
    \includegraphics[height=\toyfigheight]{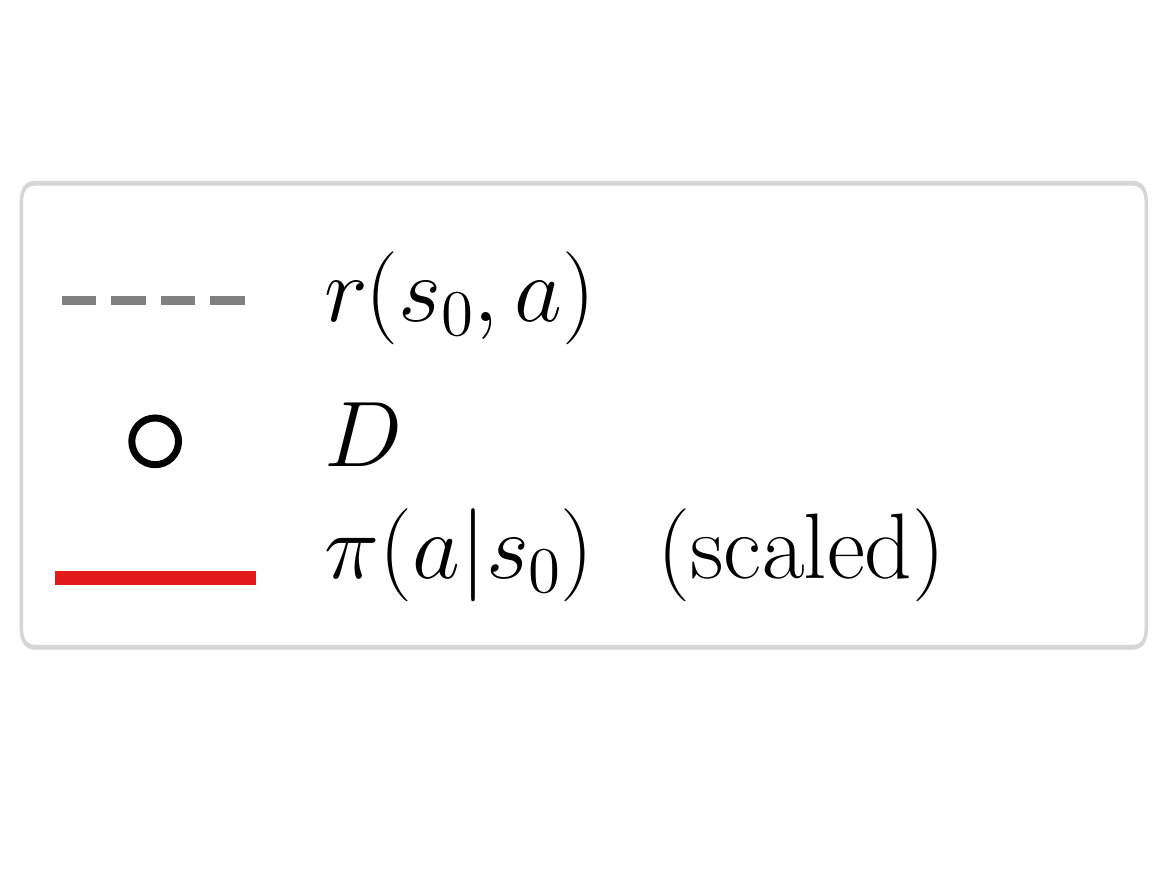}
    \vspace{-0.3in}
    \caption{Illustration on overestimation. $r(s_0, a)$ is a ground-truth reward, $D$ is a sampled offline dataset, and $\pi(a | s_0)$ is the policy density normalized by its maximum.}
    \vspace{-0.1in}
\end{figure*}

\begin{figure*}[t!]
\centering
\begin{minipage}{1.0\textwidth}
{\small \begin{align}
    &\textstyle \textbf{AlgaeDICE} ~~ 
    \big(
    e_\nu^{\textcolor{red}{\pi}}(s,a) := r(s,a) + \gamma \E_{\substack{s' \sim T(s,a), \textcolor{red}{a' \sim \pi(s')}}}[\nu(s', \textcolor{red}{a'})] - \nu(s,a), ~~ \hat e_\nu(s,a,s',a') := r(s,a) + \gamma \nu(s', a') - \nu(s,a)
    \big) \nonumber
    \\
	&\textstyle \hspace{12pt}\textcolor{red}{\max_\pi \min_\nu \max_w }
	\E_{\substack{(s,a) \sim d^D}} 
	\big[ e_\nu^\pi(s,a) w(s,a) - \alpha f\big( w(s,a) \big) \big]
	+
	(1 - \gamma) \E_{\substack{s_0 \sim p_0, a_0 \sim \pi(s_0)}} \left[ \nu(s_0, a_0) \right] \label{eq:algaedice1}
	\\
	&\textstyle = \textcolor{red}{\max_\pi \min_\nu}
	\alpha \E_{\substack{(s,a) \sim d^D}}
	\big[ f_*\big( \tfrac{1}{\alpha} e_\nu^\pi(s,a) \big) \big]
	+ (1 - \gamma) \E_{\substack{s_0 \sim p_0,  a_0 \sim \pi(s_0)}} \left[ \nu(s_0, a_0) \right] \label{eq:algaedice2}
	\\
	&\textstyle \approx \textcolor{red}{\max_\pi \min_\nu}
	\alpha \E_{\substack{(s,a,s') \sim d^D, a' \sim \pi(s')}} \big[ f_* \big( \tfrac{1}{\alpha} \hat e_\nu(s,a,s',a') \big) \big]
	+ (1 - \gamma) \E_{\substack{s_0 \sim p_0, a_0 \sim \pi(s_0)}} \left[ \nu(s_0, a_0) \right] \label{eq:algaedice3}
    \\
    &\textstyle\textbf{OptiDICE} ~~
    \big(
    e_\nu(s,a) := r(s,a) + \gamma \E_{s' \sim T(s,a)}[\nu(s')] - \nu(s), ~~ \hat e_\nu(s,a,s') := r(s,a) + \gamma \nu(s') - \nu(s),
    x_+ := \max(0, x)
    \big) \nonumber
    \\
	&\textstyle \hspace{12pt}\textcolor{blue}{\min_\nu \max_{w\ge0} }
	\E_{(s,a) \sim d^D}\big[ e_\nu(s,a) \textcolor{black}{w(s,a)} - \alpha f\big(\textcolor{black}{w(s,a)}\big) \big] + (1 - \gamma) \E_{s_0 \sim p_0}[ \nu(s_0) ] \label{eq:optidice1}
	\\
	&\textstyle = \textcolor{blue}{\min_\nu}
	\E_{(s,a) \sim d^D}\big[ e_\nu(s,a) \textcolor{black}{ (f')^{-1} \big( \tfrac{1}{\alpha} e_\nu(s,a) \big)_{+}} -\alpha f \big( \textcolor{black}{ (f')^{-1} \big( \tfrac{1}{\alpha} e_\nu(s,a) \big)_{+}} \big) \big]
	+ (1 - \gamma) \E_{s_0 \sim p_0}[ \nu(s_0) ] \label{eq:optidice2}
	\\
	&\textstyle \approx \textcolor{blue}{\min_\nu}
	\E_{(s,a,s') \sim d^D}\big[ \hat e_\nu(s,a,s') \textcolor{black}{ (f')^{-1} \big( \tfrac{1}{\alpha} \hat e_\nu(s,a,s') \big)_{+}} -\alpha f \big( \textcolor{black}{ (f')^{-1} ( \tfrac{1}{\alpha} \hat e_\nu(s,a,s') )_{+}} \big) \big]
	+ (1 - \gamma) \E_{s_0 \sim p_0}[ \nu(s_0) ] \label{eq:optidice3}
\end{align}}
\end{minipage}
\vspace{-28pt}
\end{figure*}

As discussed in Section \ref{subsection:generalization_to_gamma1}, OptiDICE can naturally be generalized to undiscounted problems ($\gamma = 1$).
In \figureref{figure:d4rl_learning_curve_gamma}, we vary $\gamma \in \{0.999, 0.9999, 1.0\}$ to validate OptiDICE's robustness in $\gamma$ by comparing with CQL in \{hopper-medium-expert, walker2d-medium-expert, halfcheetah-medium-exp\allowbreak  ert\} (See Appendix~\ref{appendix:d4rl_benchmark_gamma} for the results for other tasks).
The performance of OptiDICE stays stable, while CQL easily becomes unstable as $\gamma$ increases, due to the divergence of Q-function.
This is because OptiDICE uses 
\emph{normalized} stationary distribution corrections, whereas CQL learns the action-value function whose values becomes unbounded as $\gamma$ gets close to 1, resulting in numerical instability.

\color{green}

\section{Discussion}

Current DICE algorithms except for AlgaeDICE~\cite{nachum2019algaedice} only deal with either policy evaluation~\cite{nachum2019dualdice,zhang2019gendice,zhang2020gradientdice,zhang2020gradientdice,yang2020offpolicy,dai2020coindice} or imitation learning~\cite{kostrikov2019imitation}, not policy optimization.

Although both AlgaeDICE~\cite{nachum2019algaedice} and OptiDICE aim to solve $f$-divergence regularized RL, each algorithm solves the problem in a different way.
AlgaeDICE relies on off-policy evaluation (OPE) of the intermediate policy $\pi$ via DICE (inner $\min_\nu \max_w$ of Eq.~\eqref{eq:algaedice1}), and then optimizes $\pi$ via policy-gradient upon the OPE result (outer $\max_\pi$ of Eq.~\eqref{eq:algaedice1}), yielding an overall \textcolor{red}{$\max_\pi \min_\nu \max_w$} problem of Eq.~\eqref{eq:algaedice1}.
Although the actual AlgaeDICE implementation employs an additional approximation for practical 
optimization, i.e. using Eq.~\eqref{eq:algaedice3} that
removes the innermost $\max_w$ via convex conjugate and uses
a biased estimation of $f_*(\E_{s',a'}[\cdot])$ via $\E_{s',a'}[f_*(\cdot)]$, it still involves nested \textcolor{red}{$\max_\pi \min_\nu$} optimization, susceptible to instability.
In contrast, OptiDICE directly estimates the stationary distribution corrections of the optimal policy, resulting in \textcolor{blue}{$\min_\nu \max_w$} problem of Eq.~\eqref{eq:optidice1}.
In addition, our implementation performs the single minimization of Eq.~\eqref{eq:optidice3} (the biased estimate of \textcolor{blue}{$\min_\nu$} of \eqref{eq:optidice2}), which greatly improves the stability of overall optimization.

To see this, we conduct single-state MDP experiments, where $S = \{s_0\}$ is the state space, $A = [-1, 1]$ is the action space, $T(s_0 | s_0, a) = 1$ is the transition dynamics, $r(s_0, a)$ is a reward function, $\gamma = 0.9$, and $D$ is the offline dataset.
The blue lines in the figures present the estimates learned by each algorithm (i.e. $Q$, $\nu$, $w$) (Darker colors mean later iterations).
Similarly, the red lines visualize the action densities from intermediate policies.
In this example, a vanilla off-policy actor-critic (AC) method suffers from the divergence of Q-values
due to its TD target being outside the data distribution $d^D$.
This makes the policy learn toward unreasonably high Q-values outside $d^D$.
AlgaeDICE with Eq.~\eqref{eq:algaedice3} is no better 
for small $\alpha$.
CQL addresses this issue by lowering the Q-values outside $d^D$. 
Finally, OptiDICE computes optimal stationary distribution corrections $w(s,a) = \tfrac{d^{\pi^*}(s,a)}{d^D(s, a)}$ 
by Eq.~\eqref{eq:optidice3} and Eq.~\eqref{eq:optidice1} ($\max_w (\cdot)$ for $\nu^*$).
Then, the policy $\pi(a|s) \propto w(s, a) d^D(s, a)$ is extracted, automatically ensuring actions to be selected within the support of $d^D$ ($\mathrm{supp}(d^D)$).

Also, note that Eq.~\eqref{eq:optidice3} of OptiDICE is
\emph{unbiased} (i.e., \eqref{eq:optidice3} = \eqref{eq:optidice2}) if $T$ is deterministic (\textbf{Corollary 3}).  
In contrast, Eq.~\eqref{eq:algaedice3} of AlgaeDICE \emph{is always biased} (i.e., \eqref{eq:algaedice2} $\neq$ \eqref{eq:algaedice3}) even for the deterministic $T$, due to its dependence on expectation w.r.t. $\pi$.
Our biased objective of Eq.~\eqref{eq:optidice3} removes the
need for double sampling in Eq.~\eqref{eq:optidice2}.

\color{black}

\section{Conclusion}
We presented OptiDICE, an offline RL algorithm that aims to estimate stationary distribution corrections between the \emph{optimal} policy's stationary distribution and the dataset distribution.
We formulated the estimation problem as a minimax optimization that does not involve sampling from the target policy, which essentially circumvents the overestimation issue incurred by bootstrapped target
with out-of-distribution actions, practiced by 
most model-free offline RL algorithms.
Then, deriving the closed-form solution of the inner optimization, we simplified the nested minimax optimization for obtaining the optimal policy to a convex minimization problem.
In the experiments, we demonstrated that OptiDICE performs competitively with the state-of-the-art offline RL baselines.

\section*{Acknowledgements}
This work was supported by the National Research Foundation (NRF) of Korea (NRF-2019M3F2A1072238 and NRF-2019R1A2C1087634), and the Ministry of Science and Information communication Technology (MSIT) of Korea (IITP No. 2019-0-00075, IITP No. 2020-0-00940 and IITP No. 2017-0-01779 XAI).
We also acknowledge the support of the Natural Sciences and Engineering Research Council of Canada (NSERC) and the Canadian Institute of Advanced Research (CIFAR).

\newpage
\bibliography{main}
\bibliographystyle{icml2021}

\clearpage
\onecolumn
\icmltitle{\texorpdfstring{OptiDICE: Offline Policy Optimization via \\ Stationary Distribution Correction Estimation\\ (Supplementary Material)}{OptiDICE: Offline Policy Optimization via Stationary Distribution Correction Estimation (Supplementary Material)}}
\appendix
\section{Proof of \propositionref{proposition:closed_form_solution_v0}}
\label{appendix:proof_of_closed_form_solution_v0}

We first show that our original problem~(\ref{eq:objective_d}-\ref{eq:objective_d_constraint2}) is an instance of convex programming due to the convexity of $f$.

\begin{lemma}
The constraint optimization~(\ref{eq:objective_d}-\ref{eq:objective_d_constraint2}) is a convex optimization. 
\end{lemma}
\begin{proof}
\begin{align}
    \max_{d}~
    &
    \E_{(s, a)\sim d} [R(s, a)] - \alpha D_f(d||d^D) \tag{\ref{eq:objective_d}} \\
    \mathrm{s.t.}~
    & (\B_* d)(s) = (1 - \gamma) p_0(s) + \gamma (\T_* d)(s) ~~ \forall s, \tag{\ref{eq:objective_d_constraint1}}
    \\
	& d(s, a)\ge0 ~~ \forall s, a, \tag{\ref{eq:objective_d_constraint2}}
\end{align}
The objective function
$\E_{(s, a)\sim d} [R(s, a)] - \alpha D_f(d||d^D)$
is concave for $d:\S\times\A\rightarrow \R$ (not only for probability distribution $d\in\P(\S\times\A)$) since $D_f(d||d^D)$ is convex in $d$: 
for $t \in [0, 1]$ and any $d_1:\SA\rightarrow\R, d_2:\SA\rightarrow\R$,
\begin{align*}
    D_f((1-t)d_1+t d_2 ||d^D)
    &=
    \sum_{s, a}
    d^D(s, a)
    f
    \left(
        (1-t) \frac{d_1(s, a)}{d^D(s, a)}
        +
        t \frac{d_2(s, a)}{d^D(s, a)}
    \right)\\
    &
    <
    \sum_{s, a}
    d^D(s, a)
    \left\{
        (1-t)
        f
        \left(
            \frac{d_1(s, a)}{d^D(s, a)}
        \right)
        +
        t
        f
        \left(
             \frac{d_2(s, a)}{d^D(s, a)}
        \right)
    \right\}\\
    &=
    (1-t)D_f(d_1||d^D)
    +
    t D_f(d_2||d^D),
\end{align*}
where the strict inequality follows from assuming $f$ is strictly convex. 
In addition, the equality constraints~\eqref{eq:objective_d_constraint1} are affine in $d$, and the inequality constraints~\eqref{eq:objective_d_constraint2} are linear and thus convex in $d$.
Therefore, our problem is an instance of a convex programming, as we mentioned in Section~\ref{subsection:closed_form_solution}.
\end{proof}

In addition, by using the strong duality and the change-of-variable from $d$ to $w$, we can rearrange the original maximin optimization to the minimax optimization. 
\begin{lemma}
We assume that all states $s\in S$ are reachable 
for a given MDP.
Then,
\begin{align*}
    \max_{w\ge0}\min_\nu L(w,\nu)
    =
    \min_\nu\max_{w\ge0} L(w,\nu).
\end{align*}
\end{lemma}
\begin{proof}
    Let us define the Lagrangian of the constraint optimization~(\ref{eq:objective_d}-\ref{eq:objective_d_constraint2})
\begin{align*}
    \mathcal{L}(d, \nu, \mu)
    :=
    \E_{(s, a)\sim d} [R(s, a)] - \alpha D_f(d||d^D)
    &+
    \sum_{s}\nu(s)
    \left(
        (1-\gamma)p_0(s)
        +
        \gamma \sum_{\bar{s}, \bar{a}}T(s|\bar{s}, \bar{a})d(\bar{s}, \bar{a})
        -
        \sum_{\bar{a}}d(s, \bar{a})
    \right)\\
    &
    +\sum_{s, a}\mu(s, a)d(s, a)
\end{align*}
with Lagrange multipliers $\nu(s)~\forall s$ and $\mu(s, a)~\forall s, a$. With the Lagrangian $\mathcal{L}(d, \nu, \mu)$, the original problem~(\ref{eq:objective_d}-\ref{eq:objective_d_constraint2}) can be represented by
\begin{align*}
    \max_{d\ge0}\min_{\nu}\mathcal{L}(d, \nu, 0)
    =
    \max_{d}\min_{\nu,\mu\ge0}\mathcal{L}(d, \nu, \mu).
\end{align*}
For an MDP where 
every $s\in S$ is reachable, there always exists $d$ such that $d(s, a)>0~\forall s, a$.
From Slater's condition for convex problems (the condition that there exists 
a strictly feasible $d$~\cite{boyd2004convex}), the strong duality holds, i.e., we can change the order of optimizations:
\begin{align*}
    \max_d\min_{\nu, \mu\ge0}\mathcal{L}(d, \nu, \mu)
    =
    \min_{\nu, \mu\ge0}\max_d\mathcal{L}(d, \nu, \mu)
    =
    \min_{\nu}\max_{d\ge0}\mathcal{L}(d, \nu, 0)
    .
\end{align*}
Here, the last equality holds since 
$
    \max_{d\ge0}\mathcal{L}(d, \nu, 0)
    =
    \max_{d}\min_{\mu\ge0}\mathcal{L}(d, \nu, \mu)
    =
    \min_{\mu\ge0}\max_{d}\mathcal{L}(d, \nu, \mu)
$
for fixed $\nu$ due to the strong duality. Finally, by applying the change of variable $w=d/d^D$, we have
\begin{align*}
    \max_{w\ge0}\min_\nu L(w,\nu)
    =
    \min_\nu\max_{w\ge0} L(w,\nu).
\end{align*}

\end{proof}

Finally, the solution of the inner maximization $\max_{w\ge0} L(w,\nu)$ can be derived as follows:
\begin{customproposition}{\ref{proposition:closed_form_solution_v0}}
The closed-form solution of the inner maximization of~\eqref{eq:L_minmax}, i.e.
\begin{align*}
    w_\nu^* := \argmax_{w \ge0} (1 - \gamma) \E_{s \sim p_0} [ \nu(s) ] + \E_{(s,a) \sim d^D} \left[ - \alpha f\big(w(s,a)\big) \right] + \E_{(s,a) \sim d^D} \left[ w(s,a) \big( e_\nu(s,a) \big) \right]
\end{align*}
is given as
\begin{align}
    w_\nu^*(s, a) =
    \max \left( 0, (f')^{-1}\left(\frac{e_\nu(s, a)}{\alpha}\right) \right) ~~ \forall s, a,
\end{align}
where $(f')^{-1}$ is the inverse function of the derivative $f'$ of $f$ and is strictly increasing by strict convexity of $f$.
\end{customproposition}

\begin{proof}
For a fixed $\nu$, let the maximization $\max_{w\ge0} L(w,\nu)$ be the primal problem. Then, its corresponding dual problem is
\begin{align*}
    \max_{w}\min_{\mu\ge0}L(w, \nu)+\sum_{s, a}\mu(s, a)w(s, a).
\end{align*}
Since the strong duality holds, satisfying KKT condition is both necessary and sufficient conditions for the solutions $w^*$ and $\mu^*$ of primal and dual problems (we will use $w^*$ and $\mu^*$ instead of $w_\nu^*$ and $\mu_\nu^*$ for notational brevity).

\textbf{Condition 1 (Primal feasibility).} $w^*\ge0~\forall s, a$.

\textbf{Condition 2 (Dual feasibility).} $\mu^*\ge0~\forall s, a$.

\textbf{Condition 3 (Stationarity).} $d^D(s, a)(-\alpha f'(w^*(s, a))+e_\nu(s, a)+\mu^*(s, a))=0~\forall s, a$. 

\textbf{Condition 4 (Complementary slackness).} $w^*(s, a)\mu^*(s, a)=0~\forall s, a.$ 

From \textbf{Stationarity} and $d^D>0$, we have
\begin{align*}
	f'(w^*(s, a))=\frac{e_\nu(s, a)+\mu^*(s, a)}{\alpha}~\forall s, a
\end{align*}
and since $f'$ is invertible due to the strict convexity of $f$, 
\begin{align*}
	w^*(s, a)
	=
	(f')^{-1}
	\left(
		\frac{e_\nu(s, a)+\mu^*(s, a)}{\alpha}
	\right)~\forall s, a.
\end{align*}

Now for fixed $(s, a)\in S\times A$, let us consider two cases: either $w^*(s, a)>0$ or $w^*(s, a)=0$, where \textbf{Primal feasibility} is always satisfied in either way:

\textbf{Case 1 ($w^*(s, a)>0$).} 
$\mu^*(s, a)=0$ due to \textbf{Complementary slackness}, and thus, 
\begin{align*}
	w^*(s, a)
	=
	(f')^{-1}
	\left(
		\frac{e_\nu(s, a)}{\alpha}
	\right)>0.
\end{align*}
Note that \textbf{Dual feasibility} holds. Since $f'$ is a strictly increasing function, $e_\nu(s, a)>\alpha f'(0)$ should be satisfied if $f'(0)$ is well-defined. 

\textbf{Case 2 ($w^*(s, a)=0$).} 
$ 
	\mu^*(s, a)
	=
	\alpha f'(0)-e_\nu(s, a)\ge0
$ due to \textbf{Stationarity} and \textbf{Dual feasibility}, and thus,  $e_\nu(s, a)\le \alpha f'(0)$ should be satisfied if $f'(0)$ is well-defined. 

In summary, we have
\begin{align*}
	w_\nu^*(s, a)
	&= \max \left( 0, (f')^{-1} \left(\frac{e_\nu(s,a)}{\alpha}  \right) \right).
\end{align*}

\end{proof}

\section{Proofs of \propositionref{proposition:L_v_convex} and \textbf{Corollary}~\ref{corollary:upper_bound}}
\label{appendix:proof_of_L_v_convex}

\begin{customproposition}{\ref{proposition:L_v_convex}}
    $L(w_{\nu}^*, \nu)$ is convex with respect to $\nu$.
\end{customproposition}

\renewcommand*{\proofname}{Proof by Lagrangian duality}
\begin{proof}

Let us consider Lagrange dual function
\begin{align*}
    g(\nu,\mu)
    :=
    \max_d\mathcal{L}(d,\nu,\mu),
\end{align*}
which is always convex in Lagrange multipliers $\nu,\mu$ since $\mathcal{L}(d,\nu,\mu)$ is affine in $\nu,\mu$.
Also, for any $\mu_1, \mu_2\ge0$ and its convex combination $(1-t)\mu_1+t\mu_2$ for $0\le t \le 1$, we have 
\begin{align*}
    \min_{\mu\ge0}
    g((1-t)\nu_1+t \nu_2,\mu)
    \le
    g((1-t)\nu_1+t \nu_2,(1-t)\mu_1+t\mu_2)
    \le
    (1-t) g(\nu_1, \mu_1)
    +
    t g(\nu_2, \mu_2)
\end{align*}
by using the convexity of $g(\nu, \mu)$.
Since the above statement holds for any $\mu_1, \mu_2\ge0$, we have
\begin{align*}
    \min_{\mu\ge0}
    g((1-t)\nu_1+t \nu_2,\mu)
    \le
    (1-t) \min_{\mu_1\ge0} g(\nu_1, \mu_1)
    +
    t \min_{\mu_2\ge0}  g(\nu_2, \mu_2).
\end{align*}
Therefore, a function
\begin{align*}
    G(\nu)
    :=
    \min_{\mu\ge0}g(\nu, \mu)
    =
    \min_{\mu\ge0}\max_d\mathcal{L}(d,\nu,\mu)
    =
    \max_{d\ge0}\mathcal{L}(d, \nu, 0)
\end{align*}
is convex in $\nu$. By following the change-of-variable, we have
\begin{align*}
    \max_{d\ge0}\mathcal{L}(d, \nu, 0)
    =
    \max_{w\ge0}L(w, \nu)
    =
    L(\argmax_{w\ge0}L(w,\nu), \nu)
    =
    L(w_\nu^*, \nu)
\end{align*}
is convex in $\nu$. 

\end{proof}

\renewcommand*{\proofname}{Proof by exploiting second-order derivative}
\begin{proof}
Suppose $((f')^{-1})'$ is well-defined, where $f$ we consider in this work satisfies the condition. 
Let us define
\begin{align}
    h(x) := -f\Big(\max\Big(0, (f')^{-1}(x) \Big) \Big) + \max \Big(0,  (f')^{-1}(x) \Big)\cdot x.
    \label{eq:h_definition}
\end{align}
Then, $L(w_\nu^*, \nu)$ can be represented by using $h$:
\begin{align}
    &L(w_\nu^*, \nu) \nonumber \\
    &= (1 - \gamma) \E_{s \sim p_0}[\nu(s)] + \E_{(s,a) \sim d^D} \Big[ - \alpha f \Big( \max \Big(0,  (f')^{-1} \big( \tfrac{1}{\alpha} e_\nu(s,a) \big) \Big) \Big) + \max \Big(0,  (f')^{-1}\big( \tfrac{1}{\alpha} e_\nu(s,a) \big) \Big) e_\nu(s,a) \Big]
    \nonumber
    \\
    &= (1 - \gamma) \E_{s \sim p_0}[\nu(s)] + \E_{(s,a) \sim d^D} \Big[ \alpha h \Big( \tfrac{1}{\alpha} e_\nu(s,a) \Big) \Big] \label{eq:L_wv_h}
\end{align}
We prove that $h(x)$ is convex in $x$ by showing $h''(x) \ge 0~\forall x$.
Recall that $f'$ is a strictly increasing function by the strict convexity of $f$, which implies that $(f')^{-1}$ is also a strictly increasing function.

\textbf{Case 1.} If $(f')^{-1}(x) > 0~\forall x$,
\begin{align*}
    h(x) &= -f( (f')^{-1}(x) ) + (f')^{-1}(x) \cdot x, \\
    h'(x) &= -\underbrace{f'( (f')^{-1}(x) )}_{\text{(identity function)}} ((f')^{-1})'(x) + ((f')^{-1})'(x) \cdot x + (f')^{-1}(x) \\
          &= -x \cdot  ((f')^{-1})'(x) + ((f')^{-1})'(x) \cdot x + (f')^{-1}(x) \\
          &= (f')^{-1}(x), \\
    h''(x) &= ((f')^{-1})'(x) > 0,
\end{align*}
where $((f')^{-1})'(x) > 0$ since it is the derivative of the strictly increasing function $(f')^{-1}$.

\textbf{Case 2.} If $(f')^{-1}(x) \le 0~\forall x$,
\begin{align*}
    h(x) = -f(0) ~~\Rightarrow~~ h'(x) = 0  ~~\Rightarrow~~ h''(x) = 0.
\end{align*}
Therefore, $h''(x) \ge 0$ holds for all $x$, which implies that $h(x)$ is convex in $x$. Finally, for $t \in [0, 1]$ and any $\nu_1: S \rightarrow \R$, $\nu_2: S \rightarrow \R$,
\begin{align*}
    &L(w_{t \nu_1 + (1 - t) \nu_2}^*, t \nu_1 + (1 - t) \nu_2) \\
    &= (1 - \gamma) \E_{s \sim p_0}[ t \nu_1(s) + (1 - t) \nu_2(s) ] \\
    &~~~ + \E_{(s,a) \sim d^D} \Big[ \alpha h \Big( \tfrac{1}{\alpha} \Big( t \{R(s,a) + \gamma (\T \nu_1)(s,a) - (\B \nu_1)(s,a)\} + (1-t) \{ R(s,a) + \gamma (\T \nu_2)(s,a) - (\B \nu_2)(s,a) \} \Big) \Big) \Big] \\
    &\le t \bigg\{ (1 - \gamma) \E_{s \sim p_0}[ \nu_1(s) ] + \E_{(s,a) \sim d^D} \Big[ \alpha h\Big( \tfrac{1}{\alpha}\big( R(s,a) + \gamma (\T \nu_1)(s,a) - (\B \nu_1)(s,a) \big) \Big) \Big] \bigg\} ~~~~~~~~~~~~\text{(by convexity of $h$)}  \\
    &~~~ + (1 - t) \bigg\{ (1 - \gamma) \E_{s \sim p_0}[ \nu_2(s) ] + \E_{(s,a) \sim d^D} \Big[ \alpha h\Big( \tfrac{1}{\alpha}\big( R(s,a) + \gamma (\T \nu_2)(s,a) - (\B \nu_2)(s,a) \big) \Big) \Big] \bigg\} \\
    &= t L(w_{\nu_1}^*, \nu_1) + (1 - t) L(w_{\nu_2}^*, \nu_2)
\end{align*}
which concludes the proof.
\end{proof}

\begin{customcorollary}{\ref{corollary:upper_bound}}
$\tilde L(\nu)$ in \eqref{eq:objective_nu_biased} is an upper bound of $L(w_\nu^*, \nu)$ in \eqref{eq:objective_nu}, i.e. $L(w_\nu^*, \nu) \le \tilde L(\nu)$ always holds, where equality holds when the MDP is deterministic.
\end{customcorollary}

\renewcommand*{\proofname}{Proof by Lagrangian duality}
\begin{proof}
Let us consider a function $h$ in \eqref{eq:h_definition}.
From \propositionref{proposition:L_v_convex}, we have
$\E_{(s, a)\sim d^D}[h(\frac{1}{\alpha}e_\nu(s, a))]$ is convex in $\nu$, i.e., for $t\in[0, 1]$, $\nu_1:\S\rightarrow\R$ and $\nu_2:\S\rightarrow\R$,
\begin{align*}
    \E_{(s, a)\sim d^D}
    \left[
        h\left(\tfrac{1}{\alpha}e_{(1-t)\nu_1+t\nu_2}(s, a)\right)
    \right]
    &=
    \E_{(s, a)\sim d^D}\left[h\left(
    (1-t)\cdot \tfrac{1}{\alpha}e_{\nu_1}(s, a)
    +
    t\cdot  \tfrac{1}{\alpha}e_{\nu_2}(s, a)
    \right)\right]\\
    &\le 
    (1-t)\E_{(s, a)\sim d^D}\left[h\left(\tfrac{1}{\alpha}e_{\nu_1}(s, a)\right)\right]
    +
    t\E_{(s, a)\sim d^D}\left[h\left(\tfrac{1}{\alpha}e_{\nu_2}(s, a)\right)\right]\\
    &=
    \E_{(s, a)\sim d^D}
    \left[
        (1-t)
        \cdot
        h\left(\tfrac{1}{\alpha}e_{\nu_1}(s, a)\right)
        +
        t
        \cdot
        h\left(\tfrac{1}{\alpha}e_{\nu_2}(s, a)\right)
    \right].
\end{align*}
Since \propositionref{proposition:L_v_convex} should be satisfied for any MDP and $d^D>0$, we have
\begin{align*}
    h(
    (1-t)\cdot
    \tfrac{1}{\alpha}e_{\nu_1}(s, a)
    +
    t\cdot
    \tfrac{1}{\alpha}e_{\nu_2}(s, a)
    )
    \le 
    (1-t)\cdot
    h\left(\tfrac{1}{\alpha}e_{\nu_1}(s, a)\right)
    +
    t\cdot
    h\left(\tfrac{1}{\alpha}e_{\nu_2}(s, a)\right)
    ~\forall s, a.
\end{align*}
To prove this, if 
\begin{align*}
    h(
    (1-t)\cdot
    \tfrac{1}{\alpha}e_{\nu_1}(s, a)
    +
    t\cdot
    \tfrac{1}{\alpha}e_{\nu_2}(s, a)
    )
    >
    (1-t)\cdot
    h\left(\tfrac{1}{\alpha}e_{\nu_1}(s, a)\right)
    +
    t\cdot
    h\left(\tfrac{1}{\alpha}e_{\nu_2}(s, a)\right)
    ~\exists s, a,
\end{align*}
we can always find out $d^D>0$ that contradicts \propositionref{proposition:L_v_convex}.
Also, since $\tfrac{1}{\alpha}e_{\nu}(s, a)$ can have an arbitrary real value, $h$ should be a convex function. Therefore, it can be shown that
\begin{align*}
    h\left(
        \E_{s'\sim T(s, a)}
        \left[
            \tfrac{1}{\alpha}\hat{e}_\nu(s, a, s')
        \right]
    \right)
    \le
    \E_{s'\sim T(s, a)}
        \left[
            h\left(\tfrac{1}{\alpha}\hat{e}_\nu(s, a, s')\right)
        \right]
    ~\forall s, a,
\end{align*}
due to Jensen's inequality, and thus,
\begin{align*}
    \E_{(s, a)\sim d^D}
    \left[ 
        h\left(\tfrac{1}{\alpha}e_\nu(s, a)\right)
    \right]
    =
    \E_{(s, a)\sim d^D}
    \left[ 
        h
        \left(
            \E_{s'\sim T(s, a)}
            \left[
                \tfrac{1}{\alpha}\hat{e}_\nu(s, a, s')
            \right]
        \right)
    \right]
    \le
    \E_{(s, a, s')\sim d^D}
    \left[
        h\left(
            \tfrac{1}{\alpha}\hat{e}_\nu(s, a, s')
        \right)
    \right].
\end{align*}
Also, the inequality becomes tight when the transition model is deterministic since $h \big(\tfrac{1}{\alpha} \E_{s' \sim T(s,a)}[\hat e(s, a,s')] \big) = \E_{s' \sim T(s,a)}[h \big( \tfrac{1}{\alpha} \hat e(s, a, s')\big) ]$ should always hold for the deterministic transition $T$.
\end{proof}

\renewcommand*{\proofname}{Proof by exploiting second-order derivative}
\begin{proof}

We start from \eqref{eq:L_wv_h} in the proof of \propositionref{proposition:L_v_convex}.
\begin{align}
    L(w_\nu^*, \nu)
    &= (1 - \gamma) \E_{s \sim p_0}[\nu(s)] + \E_{(s,a) \sim d^D} \Big[ \alpha h \Big( \tfrac{1}{\alpha} e_\nu(s,a) \Big) \Big] \tag{\ref{eq:L_wv_h}} \\
    &=(1 - \gamma) \E_{s \sim p_0}[\nu(s)] + \E_{(s,a) \sim d^D} \Big[ \alpha h \Big( \tfrac{1}{\alpha} \E_{s' \sim T(s,a)}[ \hat e_\nu(s,a,s') ] \Big) \Big] \nonumber \\
    &\le (1 - \gamma) \E_{s \sim p_0}[\nu(s)] + \E_{\substack{(s,a) \sim d^D \\ s' \sim T(s,a)}} \Big[ \alpha h \Big( \tfrac{1}{\alpha} \hat e_\nu(s,a,s') \Big) \Big] \hspace{20pt} \text{(by Jensen's inequality with the convexity of $h$)} \nonumber \\
    &= (1 - \gamma) \E_{s \sim p_0}[\nu(s)] +\E_{(s,a,s') \sim d^D}\bigg[-  \alpha f \Big( \max \Big(0,  (f')^{-1} \big( \tfrac{1}{\alpha} \hat e_\nu(s,a,s') \big) \Big) \Big) \hspace{38pt} \text{(by definition of $h$)} \nonumber \\
    &\hspace{150pt}+ \max \Big(0,  (f')^{-1}\big( \tfrac{1}{\alpha}  \hat e_\nu(s,a,s') \big) \Big) \big( \hat e_\nu(s,a,s') \big) \bigg] = \tilde L(\nu) \tag{\ref{eq:objective_nu_biased}}
\end{align}
Also, Jensen's inequality becomes tight when the transition model is deterministic for the same reason we describe in \textit{Proof by Lagrangian duality}.
\end{proof}

\section{OptiDICE for Finite MDPs}
\label{appendix:optidice_for_finite_mdp}

For tabular MDP experiments, we assume that the data-collection policy is given to OptiDICE for a fair comparison with SPIBB~\cite{laroche2019safe} and BOPAH~\cite{lee2020batch}, which directly exploit the data-collection policy $\piD$. However, the extension of tabular OptiDICE to not assuming $\piD$ is straightforward.

As a first step, we construct an MLE MDP $\hat M = \langle S, A, T, R, p_0, \gamma \rangle$ using the given offline dataset. 
Then, we compute a stationary distribution of the data-collection policy $\piD$ on the MLE MDP, denoted as $d^{\piD}$. Finally, we aim to solve the following policy optimization problem on the MLE MDP:
\begin{align*}
    \pi^* := \argmax_{\pi} \E_{(s, a)\sim d^\pi}[R(s, a)] - \alpha D_f(d^\pi||d^{\piD}),
\end{align*}
which can be reformulated in terms of optimizing the stationary distribution corrections $w$ with Lagrange multipliers $\nu$:
\begin{align}
    \min_\nu \max_{w \ge 0} L(w, \nu) 
    &= (1 - \gamma) \E_{s \sim p_0(s)} \left[ \nu(s) \right] + \E_{(s,a) \sim \D} \bigg[ - \alpha f\big( w(s,a) \big) + w(s,a) \Big( R(s,a)  + \gamma (\T \nu)(s,a) - (\B \nu)(s, a) \Big) \bigg]. \label{eq:tabular_optidice_objective_minimax}
\end{align}
For tabular MDPs, we can describe the problem using vector-matrix notation.
Specifically, $\nu \in \R^{|S|}$ is represented as a $|S|$-dimensional vector, $w \in \R^{|S||A|}$ by $|S||A|$-dimensional vector, and $R \in \R^{|S||A|}$ by $|S||A|$-dimensional reward vector.
Then, we denote $D = \mathrm{diag}(d^{\piD}) \in \R^{|S||A| \times |S||A|}$ as a diagonal matrix, $\T \in \R^{|S||A| \times |S|}$ as a matrix, and $\B \in \R^{|S||A| \times |S|}$ as a matrix that satisfies
\begin{align*}
    \T \nu \in \R^{|S||A|} ~~~&\mathrm{s.t.}~~~ (\T \nu)((s,a)) = \sum_{s'} T(s'|s,a) \nu(s') \\
    \B \nu \in \R^{|S||A|} ~~~&\mathrm{s.t.}~~~ (\B \nu)((s,a)) =  \nu(s)
\end{align*}
For brevity, we only consider the case where $f(x) = \frac{1}{2}(x - 1)^2$ that corresponds to $\chi^2$-divergence-regularized policy optimization, and the problem~\eqref{eq:tabular_optidice_objective_minimax} becomes
\begin{align}
    \min_\nu \max_{w \ge 0} L(w, \nu) 
    &= (1 - \gamma) p_0^\top v - \frac{\alpha}{2} (w - 1)^\top D (w - 1) + w^\top D (R + \gamma \T \nu - \B v) 
\end{align}
From \propositionref{proposition:closed_form_solution_v0}, we have the closed-form solution of the inner maximization as $w_\nu^* = \max\big(0, \frac{1}{\alpha}(R + \gamma \T \nu - \B \nu) + 1\big)$ since $(f')^{-1}(x) = x + 1$. By plugging $w_\nu^*$ into $L(w, \nu)$, we obtain
\begin{align}
    \min_\nu L(w_\nu^*, \nu) = L(\nu)
    &:= (1 - \gamma) p_0^\top \nu - \frac{\alpha}{2} (w_\nu^* - 1) D (w_\nu^* - 1) + w_\nu^{*\top} D (R + \gamma \T \nu - \B v)
    \label{eq:tabular_optidice_objective_min}
\end{align}
Since $L(\nu)$ is convex in $\nu$ by \propositionref{proposition:L_v_convex}, we perform a second-order optimization, i.e., Newton's method, to compute an optimal $\nu^*$ efficiently. For almost every $\nu$, we can compute the first and second derivatives as follows:
\begin{align*}
    e_\nu :=& R + \gamma \T \nu - \B \nu & \text{(advantage using $\nu$)} \\
    m :=& \1 \left(\tfrac{1}{\alpha} e_\nu + 1 \ge 0\right) & \text{(binary masking vector)} \\
    w_\nu^* :=& \big(\tfrac{1}{\alpha} e_\nu + 1\big) \odot m \hspace{32pt} \text{(where $\odot m$ denotes element-wise masking)} &\text{(closed-form solution)} \\
    J :=& \frac{\partial w_\nu^*}{\partial \nu} = \tfrac{1}{\alpha}(\gamma \T - \B) \odot m \hspace{10pt} \text{(where $\odot m$ denotes row-wise masking)} & \text{(Jacobian matrix)}\\
    g :=& \frac{\partial L(\nu)}{\partial \nu}
    = (1 - \gamma) p_0 - \alpha J^\top D (w_\nu^* - 1) + J^\top D e_\nu + (\gamma \T - \B)^{\top} D w_\nu^*  & \text{(first-order derivative)} \\
    H :=& \frac{\partial^2 L(\nu)}{\partial \nu^2}
    = -\alpha J^\top D J + J^\top D (\gamma \T - \B) + (\gamma \T - \B)^\top D J & \text{(second-order derivative)}.
\end{align*}
We iteratively update $\nu$ in the direction of $-H^{-1} g$ until convergence. Finally, $w_{\nu^*}^*$ and the corresponding optimal policy $\pi^*(a|s) \propto w_{\nu^*}^*(s,a) \cdot d^{\piD}(s,a)$ are computed.
The pseudo-code of these procedures is presented in \textbf{Algorithm~\ref{alg:tabular_optidice}}.

\begin{algorithm}[h!]
\caption{Tabular OptiDICE ($f(x) = \frac{1}{2} (x - 1)^2$)}
\centering
\begin{algorithmic}[0]
	\REQUIRE MLE MDP $\hat M=\langle S, A, \T, r, \gamma, p_0 \rangle$, data-collection policy $\piD$, regularization hyperparameter $\alpha > 0$.
	\STATE $d^{\piD} \leftarrow \textsc{ComputeStationaryDistribution}(\hat M, \piD)$
	\STATE $D \leftarrow \mathrm{diag}(d^\piD)$
	\STATE $\nu \leftarrow \text{(random initialization)}$
	\WHILE{$\nu$ is not converged}
		\STATE $e \leftarrow r + \gamma \T v - \B v$
		\STATE $m \leftarrow \1 \big( \tfrac{1}{\alpha} e_\nu + 1 \ge 0 \big)$
		\STATE $w \leftarrow \big(\frac{1}{\alpha} e + 1 \big) \odot m $
		\STATE $J \leftarrow \frac{1}{\alpha} \big( \gamma \T - \B  \big) \odot m$
		\STATE $g \leftarrow (1 - \gamma) p_0 - \alpha J^\top D (w - 1) + J^\top D e + (\gamma \T - \B)^{\top} D w$
		\STATE $H \leftarrow -\alpha J^\top D J + J^\top D (\gamma \T - \B) + (\gamma \T - \B)^\top D J$
		\STATE $\nu \leftarrow \nu - \eta H^{-1} g$  ~ (where $\eta$ is a step-size)
	\ENDWHILE
    \begin{flalign}
    &\begin{aligned}
        \pi^*(a | s) \leftarrow \frac{w(s,a) d^{\piD}(s,a)}{\sum_{a'} w(s,a') d^{\piD}(s,a')} ~~~ \forall s,a
    \end{aligned}&& \nonumber
    \end{flalign}
	\ENSURE $\pi^*, w$
\end{algorithmic}
\label{alg:tabular_optidice}
\end{algorithm}

\newpage
\section{Proof of \propositionref{proposition:closed_form_solution_v1}}
\label{appendix:proof_of_closed_form_solution_v1}

\begin{customproposition}{\ref{proposition:closed_form_solution_v1}}
The closed-form solution of the inner maximization with normalization constraint, i.e., 
\begin{align*}
    w_{\nu,\lambda}^* := \argmax_{w \ge0} (1 - \gamma) \E_{s \sim p_0} [ \nu(s) ] + \E_{(s,a) \sim d^D} \left[ - \alpha f\big(w(s,a)\big) \right] + \E_{(s,a) \sim d^D} \left[ w(s,a) \big( e_\nu(s,a) -\lambda \big) \right]
    +
    \lambda
\end{align*}
is given as
\begin{align*}
    w_{\nu,\lambda}^*(s, a)
    =
    \max
    \left(
        0, (f')^{-1}\left(\frac{e_\nu(s, a)-\lambda}{\alpha}\right)
    \right).
\end{align*}
\end{customproposition}

\renewcommand*{\proofname}{Proof}
\begin{proof}
Similar to the proof for \propositionref{proposition:closed_form_solution_v0}, we consider the maximization problem 
\begin{align*}
    \max_{w\ge0} L(w,\nu,\lambda)
\end{align*}
for fixed $\nu$ and $\lambda$, where we consider this maximization as a primal problem. Then, its dual problem is
\begin{align*}
    \max_{w}\min_{\mu\ge0}L(w, \nu,\lambda)+\sum_{s, a}\mu(s, a)w(s, a).
\end{align*}
Since the strong duality holds, KKT condition is both necessary and sufficient conditions for primal and dual solutions $w^*$ and $\mu^*$, where dependencies on $\nu,\lambda$ are ignored for brevity. 
While the KKT conditions on \textbf{Primal feasibility}, \textbf{Dual feasibility} and \textbf{Complementary slackness} are the same as those in the proof of \propositionref{proposition:closed_form_solution_v0},
the condition on \textbf{Stationarity} is slighted different due to the normalization constraint:

\textbf{Condition 1 (Primal feasibility).} $w^*\ge0~\forall s, a$.

\textbf{Condition 2 (Dual feasibility).} $\mu^*\ge0~\forall s, a$.

\textbf{Condition 3 (Stationarity).} $d^D(s, a)(-\alpha f'(w^*(s, a))+e_\nu(s, a)+\mu^*(s, a)-\lambda)=0~\forall s, a$. 

\textbf{Condition 4 (Complementary slackness).} $w^*(s, a)\mu^*(s, a)=0~\forall s, a.$ 

The remainder of the proof is similar to the proof of \propositionref{proposition:closed_form_solution_v0}. 
From \textbf{Stationarity} and $d^D>0$, we have
\begin{align*}
	f'(w^*(s, a))=\frac{e_\nu(s, a)+\mu^*(s, a)-\lambda}{\alpha}~\forall s, a,
\end{align*}
and since $f'$ is invertible due to the strict convexity of $f$, 
\begin{align*}
	w^*(s, a)
	=
	(f')^{-1}
	\left(
		\frac{e_\nu(s, a)+\mu^*(s, a)-\lambda}{\alpha}
	\right)~\forall s, a.
\end{align*}
Given $s, a$, assume either $w^*(s, a)>0$ or $w^*(s, a)=0$, satisfying \textbf{Primal feasibility}.

\textbf{Case 1 ($w^*(s, a)>0$).} 
$\mu^*(s, a)=0$ due to \textbf{Complementary slackness}, and thus, 
\begin{align*}
	w^*(s, a)
	=
	(f')^{-1}
	\left(
		\frac{e_\nu(s, a)-\lambda}{\alpha}
	\right)>0.
\end{align*}
Note that \textbf{Dual feasibility} holds. Since $f'$ is a strictly increasing function due to the strict convexity of $f$, $e_\nu(s, a)-\lambda>\alpha f'(0)$ should be satisfied if $f'(0)$ is well-defined. 

\textbf{Case 2 ($w^*(s, a)=0$).} 
$ 
	\mu^*(s, a)
	=
	\alpha f'(0)-e_\nu(s, a)+\lambda\ge0
$ due to \textbf{Stationarity} and \textbf{Dual feasibility}, and thus,  $e_\nu(s, a)-\lambda\le \alpha f'(0)$ should be satisfied if $f'(0)$ is well-defined. 

In summary, we have
\begin{align*}
	w_{\nu,\lambda}^*(s, a)
	&= \max \left( 0, (f')^{-1} \left(\frac{e_\nu(s,a)-\lambda}{\alpha}  \right) \right).
\end{align*}
\end{proof}

\section{$f$-divergence}
\label{appendix:f_divergence}

\begin{figure}[h!]
    \vspace{-0.1in}
    \centering 
    ~~
    \subfigure[]{\includegraphics[width=0.25\columnwidth]{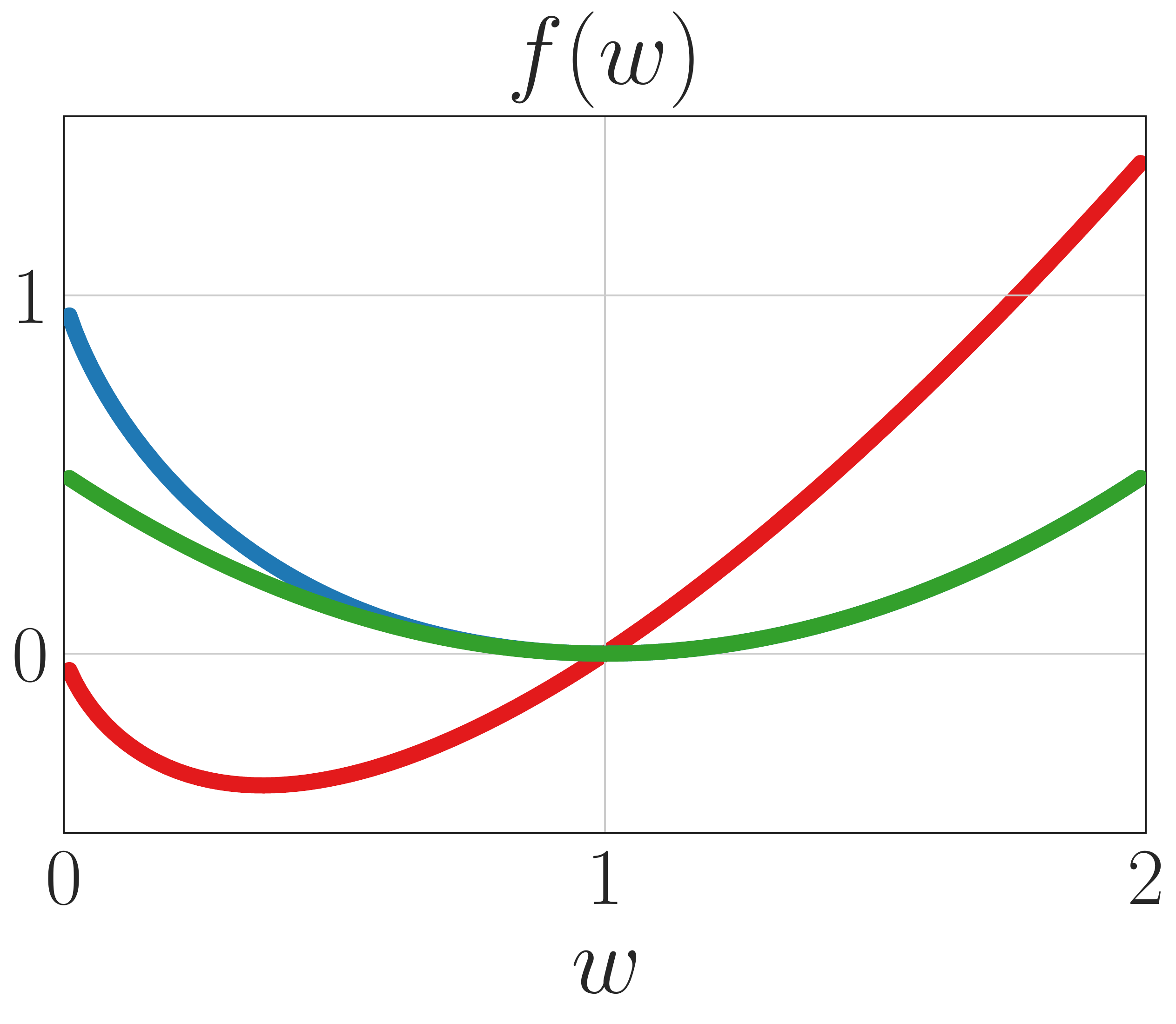}~~~}
     \subfigure[]{\includegraphics[width=0.25\columnwidth]{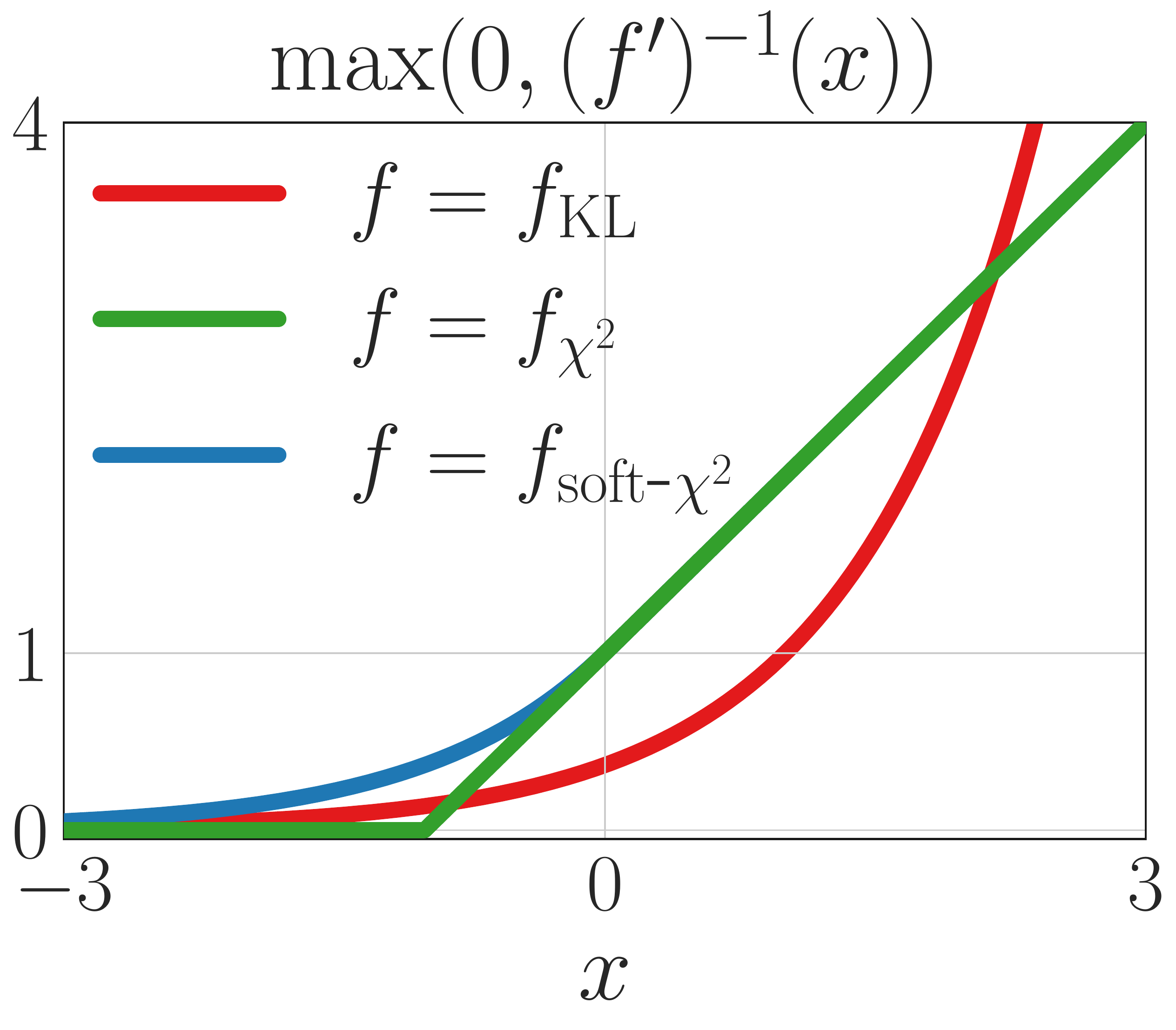}~~~}
	\caption{We depict (a) generator functions $f$ of $f$-divergences and (b) corresponding functions $\max(0, (f')^{-1}(\cdot))$ used to define the closed-form solution in \propositionref{proposition:closed_form_solution_v1}.
	While $f_{\mathrm{KL}}(x)$ has a numerical instability for large $x$ and $f_{\chi^2}(x)$ provides zero gradients for negative $x$, 
	$f_{\mathrm{soft}\text{-}\chi^2}$ does not suffer from both issues.
	}
	\label{figure:f_divergence}
\end{figure}


Pertinent to the result of \propositionref{proposition:closed_form_solution_v1}, one can observe that the choice of the function $f$ of $f$-divergence can affect the numerical stability of optimization especially when using the closed-form solution of $w_{\nu,\lambda}^*$:
\begin{align*}
    w_{\nu,\lambda}^*(s, a) =
    \max \left( 0, (f')^{-1}\left(\frac{e_\nu(s, a)-\lambda}{\alpha}\right) \right).
\end{align*}

For example, for the choice of $f(x) = f_\mathrm{KL}(x) := x \log x$ that corresponds to KL-divergence, we have $(f_\mathrm{KL}')^{-1}(x) = \exp(x - 1)$.
This yields the following closed-form solution of $w_{\nu,\lambda}^*$:
\begin{align*}
    w_{\nu,\lambda}^*(s, a) =
    \exp\left(\frac{e_\nu(s, a)-\lambda}{\alpha} - 1 \right).
\end{align*}
However, the choice of $f_\mathrm{KL}$ can incur numerical instability due to its inclusion of an $\exp(\cdot)$, i.e. for values of $\frac{1}{\alpha} (e_\nu(s,a) - \lambda)$ in order of tens, the value of $w_{\nu, \lambda}^*(s,a)$ easily explodes and so does the gradient $\nabla_\nu w_{\nu, \lambda}^*(s,a)$.

Alternatively, for the choice of $f(x) = f_{\chi^2}(x) := \frac{1}{2}(x-1)^2$ that corresponds to $\chi^2$-divergence, we have $(f_{\chi^2}')^{-1}(x) = x + 1$. This yields the following closed-form solution of $w_{\nu, \lambda}^*$:
\begin{align*}
    w_{\nu,\lambda}^*(s, a)
    =
    \mathrm{ReLU}
    \left(\frac{e_\nu(s, a)-\lambda}{\alpha} + 1\right),
\end{align*}
where $\mathrm{ReLU}(x) := \max(0, x)$. Still, this choice may suffer from dying gradient problem: for values of negative $\frac{1}{\alpha} (e_\nu(s,a) - \lambda) + 1$,
the gradient $\nabla_\nu w_{\nu, \lambda}^*(s,a)$ becomes zero, which can make training $\nu$ slow or even fail.

Consequently, we adopt the function $f = f_{\mathrm{soft}\text{-}\chi^2}$ that combines the form of $f_\mathrm{KL}$ and $f_{\chi^2}$, which can prevent both of the aforementioned issues:
\begin{align*}
    f_{\mathrm{soft}\text{-}\chi^2}(x) :=
	\begin{cases}
		x\log x-x+1	&\text{if } 0<x<1\\
		\frac{1}{2} (x - 1)^2 &\text{if }x\ge1.
	\end{cases} 
	~~~ \Rightarrow ~~~
	(f_{\mathrm{soft}\text{-}\chi^2}(x)')^{-1}(x) =
	\begin{cases}
	    \exp(x) & \text{if } x < 0 \\
	    x + 1 & \text{if } x \ge 0
	\end{cases}
\end{align*}
This particular choice of $f$ yields the following closed-form solution of $w_{\nu,\lambda}^*$:
\begin{align*}
    w_{v,\lambda}^*(s, a)
    =
    \mathrm{ELU}
    \left(\frac{e_v(s, a)-\lambda}{\alpha}\right)+1.
\end{align*}
Here, $\mathrm{ELU}(x):=\exp(x)-1$ if $x<0$ and $x$ for $x\ge0$. 
Note that 
the solution for $f=f_{\mathrm{soft}\text{-}\chi^2}$ is numerically 
stable for large $\frac{1}{\alpha} (e_\nu(s,a) - \lambda)$ and always gives non-zero gradients.
We use $f=f_{\mathrm{soft}\text{-}\chi^2}$ for the D4RL experiments.

\section{Experimental Settings}

\subsection{Random MDPs}
\label{appendix:random_mdps}
We validate tabular OptiDICE's efficiency and robustness using randomly generated MDPs with varying numbers of trajectories and the degree of optimality of the data-collection policy, where we follow the experimental protocol of \cite{laroche2019safe,lee2020batch}.
We conduct repeated experiments for 10,000 runs. For each run, an MDP is generated randomly, and a data-collection policy is constructed according to the given degree of optimality $\zeta \in \{0.9, 0.5\}$.
Then, $N$ trajectories for $N \in \{10, 20, 50, 100, 200, 500, 1000, 2000\}$ are collected using the generated MDP and the data-collection policy $\piD$.
Finally, the constructed data-collection policy and the collected trajectories are given to each offline RL algorithm, and we measure the mean performance and the CVaR 5\% performance.

\subsubsection{Random MDP generation}
We generate random MDPs with $|S| = 50$, $|A| = 4$, $\gamma = 0.95$, and a deterministic initial state distribution, i.e. $p_0(s) = 1$ for a fixed $s = s_0$.
The transition model has connectivity 4: for each $(s,a)$, non-zero probabilities of transition to next states are given to four different states $(s_1', s_2', s_3', s_4')$, where the random transition probabilities are sampled from a Dirichlet distribution $[p(s_1' | s,a), p(s_2' | s,a), p(s_3' | s,a), p(s_4' | s,a)] \sim \mathrm{Dir}(1,1,1,1)$.
The reward of 1 is given to one state that minimizes the optimal state value at the initial state; other states have zero rewards.
This design of the reward function can be understood as we choose a goal state that is the most difficult to reach from the initial state.
Once the agent reaches the rewarding goal state, the episode terminates.

\subsubsection{Data-collection policy construction}

The notion of $\zeta$-optimality of a policy is defined as a relative performance with respect to a uniform random policy $\pi_\mathrm{unif}$ and an optimal policy $\pi^*$:
\begin{align*}
    \text{\big($\zeta$-optimal policy $\pi$'s performance $V^{\pi}(s_0)$\big)} 
    = \zeta V^{*}(s_0) + (1 - \zeta) V^{\pi_\mathrm{unif}}(s_0)
\end{align*}
However, there are infinitely many ways to construct a $\zeta$-optimal policy.
In this work, we follow the way introduced in \citet{laroche2019safe} to construct a $\zeta$-optimal data-collection policy, and the process proceeds as follows.
First, an optimal policy $\pi^*$ and the optimal value function $Q^*$ are computed.
Then, starting from $\pi_\mathrm{soft} := \pi^*$, the policy $\pi_\mathrm{soft}$ is softened via $\pi_\mathrm{soft} \propto \exp(Q^*(s,a) / \tau)$ by increasing the temperature $\tau$ until the performance reaches $\frac{\zeta + 1}{2}$-optimality.
Finally, the softened policy $\pi_\mathrm{soft}$ is perturbed by discounting action selection probability of an optimal action at randomly selected state.
This perturbation continues until the performance of the perturbed policy reaches $\zeta$-optimality.
The pseudo-code for the process of the data-collection policy construction is presented in \textbf{Algorithm~\ref{alg:data_policy_generation}}.

\begin{algorithm}[h!]
   \caption{Data-collection policy construction}
\begin{algorithmic}
   \STATE {\bfseries Input:} MDP $M$, Degree of optimality of the data-collection policy $\zeta$
   \STATE Compute the optimal policy $\pi^*$ and its value function $Q^*(s,a)$ on the given MDP $M$.
   \STATE Initialize $\pi_\mathrm{soft} \leftarrow \pi^*$
   \STATE Initialize a temperature parameter $\tau \leftarrow 10^{-7}$
   \WHILE{$V^{\pi_\mathrm{soft}}(s_0) > \frac{1}{2} V^*(s_0) + \frac{1}{2}
   \Big( \zeta V^*(s_0) + (1 - \zeta) V^{\pi_\mathrm{unif}}(s_0) \Big)$}
        \STATE Set $\pi_{\mathrm{soft}}$ to $\pi_{\mathrm{soft}}(a | s) \propto \exp \left( \frac{Q^*(s,a)}{\tau} \right)$ ~ $\forall s,a$
        \STATE $\tau \leftarrow \tau / 0.9$
   \ENDWHILE
   \STATE Initialize $\piD \leftarrow \pi_\mathrm{soft}$
   \WHILE{$V^{\piD}(s_0) > \zeta V^*(s_0) + (1 - \zeta) V^{\pi_\mathrm{unif}}(s_0) $}
        \STATE Sample $s \in S$ uniformly at random.
        \STATE $\piD(a^* | s) \leftarrow 0.9 \piD(a^* | s)$ where $a^* = \argmax_a Q^*(s,a)$.
        \STATE Normalize $\piD(\cdot | s)$ to ensure $\sum_{a} \piD(a|s) = 1$.
   \ENDWHILE
   \STATE {\bfseries Output:} The data-collection policy $\piD$
\end{algorithmic}
\label{alg:data_policy_generation}
\end{algorithm}

\newpage
\subsubsection{Hyperparameters}
We compare our tabular OptiDICE with BasicRL, RaMDP~\cite{petrik2016safe}, RobustMDP~\cite{Nilim2005robust,Iyengar2005robust}, SPIBB~\cite{laroche2019safe}, and BOPAH~\cite{lee2020batch}.
For the hyperparameters, we follow the setting in the public code of SPIBB and BOPAH, which are listed as follows:

\textbf{RaMDP.} $\kappa=0.003$ is used for the reward-adjusting hyperparameter.

\textbf{RobustMDP.} $\delta = 0.001$ is used for the confidence interval hyperparameter to construct an uncertainty set.

\textbf{SPIBB.} $N_{\wedge}=5$ is used for the data-collection policy bootstrapping threshold.

\textbf{BOPAH.} The 2-fold cross validation criteria and fully state-dependent KL-regularization is used.

\textbf{OptiDICE.} $\alpha = N^{-1}$ for the number $N$ of trajectories is used for the reward-regularization balancing hyperparameter. We also use $f(x) = \frac{1}{2}(x-1)^2$ which corresponds to $\chi^2$-divergence.


\subsection{D4RL benchmark}
\label{appendix:d4rl_benchmark}

\subsubsection{Task descriptions}
We use Maze2D and Gym-MuJoCo environments of D4RL benchmark~\cite{fu2020d4rl} to evaluate OptiDICE and CQL~\cite{kumar2020conservative} in continuous control tasks.
We summarize the descriptions of tasks in D4RL paper~\cite{fu2020d4rl} as follows: 

\textbf{Maze2D.}
This is a navigation task in 2D state space, while the agent tries to reach a fixed goal location. 
By using priorly gathered trajectories, the goal of the agent is to find out a shortest path to reach the goal location. 
The complexity of the maze increases with the order of "maze2d-umaze", "maze2d-medium" and "maze2d-large".

\textbf{Gym-MuJoCo.}
For each task in $\{$hopper, walker2d, halfcheetah$\}$ of MuJoCo continuous controls, the dataset is gathered in the following ways. 

\textit{random.}
The dataset is generated by a randomly initialized policy in each task. 

\textit{medium.}
The dataset is generated by using the policy trained by SAC~\cite{haarnoja2018soft} with early stopping.

\textit{medium-replay.}
The ``replay" dataset consists of the samples gathered during training the policy for ``medium" dataset. The ``medium-replay" dataset includes both ``medium" and ``replay" datasets.

\textit{medium-expert.} 
The dataset is given by using the same amount of expert trajectories and suboptimal trajectories, where those suboptimal ones are gathered by using  either a randomly uniform policy or a medium-performance policy.

\subsubsection{Hyperparameter settings for CQL}

We follow the hyperparameters specified by~\citet{kumar2020conservative}. For learning both the Q-funtions and the policy, fully-connected multi-layer perceptrons (MLPs) with three hidden layers and ReLU activations are used, where the number of hidden units on each layer is equal to 256. A Q-function learning rate of 0.0003 and a policy learning rate of 0.0001 are used with Adam optimizer for these networks. 
CQL$(\mathcal{H})$ is evaluated, with an approximate max-backup (see Appendix F of~\citep{kumar2020conservative} for more details) and a static $\alpha=5.0$, which controls the conservativeness of CQL.
The policy of CQL is updated for 2,500,000 iterations, while we use 40,000 warm-up iterations where we update Q-functions as usual, but the policy is updated according to the behavior cloning objective.

\subsubsection{Hyperparameter settings for OptiDICE}
\label{hyperparams}

For neural networks $\nu_\theta, e_\phi, \pi_\psi$ and $\pi_\beta$ in \textbf{Algorithm~\ref{alg:optidice}},
we use fully-connected MLPs with two hidden layers and ReLU activations, where the number of hidden units on each layer is equal to 256.  
For $\pi_\psi$, we use tanh-squashed normal distribution.
We regularize the entropy of $\pi_\psi$ with learnable entropy regularization coefficients, where target entropies are set to be the same as those in SAC~\cite{haarnoja2018soft} ($-\mathrm{dim}(\A)$ for each task). 
For $\pi_\beta$, we use tanh-squashed mixture of normal distributions, where we build means and standard deviations of each mixture component upon shared hidden outputs. 
No entropy regularization is applied to $\pi_\beta$. 
For both $\pi_\psi$ and $\pi_\beta$, means are clipped within $(-7.24, 7.24)$, while log of standard deviations are clipped within $(-5, 2)$.
For the optimization of each network, we use stochastic gradient descent with Adam optimizer and its learning rate $0.0003$. 
The batch size is set to be $512$. 
Before training neural networks, we preprocess the dataset $D$ by standardizing observations and rewards.
We additionally scale the rewards by multiplying 0.1.
We update the policy $\pi_\psi$ for 2,500,000 iterations, while we use 500,000 warm-up iterations for other networks, i.e., those networks other than $\pi_\psi$ are updated for 3,000,000 iterations.

For each task and OptiDICE methods (OptiDICE-minimax, OptiDICE-MSE), we search the number $K$ of mixtures (for $\pi_\beta$) within $\{1, 5, 9\}$ and the coefficient $\alpha$ within $\{0.0001, 0.001, 0.01, 0.1, 1\}$, while we additionally search $\alpha$ over $\{2, 5, 10\}$ for hopper-medium-replay. The hyperparameters $K$ and $\alpha$ showing the best mean performance were chosen, which are described as follows:

\begin{table}[h!]
\centering
\caption{Hyperaparameters}
\begin{tabular}{l | r r | r r}
\toprule
\multirow{2}{*}{Task} & \multicolumn{2}{c}{OptiDICE-MSE} & \multicolumn{2}{c}{OptiDICE-minimax} \\ \cline{2-5}
& $K$ & $\alpha$ & $K$ & $\alpha$ \\
\midrule
maze2d-umaze                    & 5         & 0.001         & 1         & 0.01          \\
maze2d-medium                   & 5         & 0.0001        & 1         & 0.01          \\
maze2d-large                    & 1         & 0.01          & 1         & 0.01          \\
\midrule
hopper-random                   & 5         & 1             & 5         & 1             \\
hopper-medium                   & 9         & 0.1           & 9         & 0.1           \\
hopper-medium-replay            & 9         & 10            & 1         & 2             \\
hopper-medium-expert            & 9         & 1             & 5         & 1             \\
\midrule
walker2d-random                 & 9         & 0.0001        & 1         & 0.0001        \\
walker2d-medium                 & 9         & 0.01          & 5         & 0.01          \\
walker2d-medium-replay          & 9         & 0.1           & 9         & 0.1           \\
walker2d-medium-expert          & 5         & 0.01          & 5         & 0.01          \\
\midrule
halfcheetah-random              & 5         & 0.0001        & 9         & 0.001         \\
halfcheetah-medium              & 1         & 0.01          & 1         & 0.1           \\
halfcheetah-medium-replay       & 9         & 0.01          & 1         & 0.1           \\
halfcheetah-medium-expert       & 9         & 0.01          & 9         & 0.01          \\
\bottomrule
\end{tabular}
\end{table}

\newpage
\section{Experimental results}
\label{appendix:d4rl_benchmark_gamma}

\subsection{Experimental results for $\gamma=0.99$}
\begin{table*}[!h]
\caption{Normalized performance of OptiDICE compared with baselines. Mean scores for baselines---BEAR~\cite{kumar2019stabilizing}, BRAC~\cite{wu2019behavior}, AlgaeDICE~\cite{nachum2019algaedice}, and CQL~\cite{kumar2020conservative}--- come from D4RL benchmark. 
We also report the performance of CQL~\cite{kumar2020conservative} obtained by running the code released by authors (denoted as CQL (ours) in the table).
OptiDICE achieves the best performance on 6 tasks compared to our baselines.
Note that 3-run mean scores without confidence intervals were reported on each task by~\citet{fu2020d4rl}. For CQL (ours) and OptiDICE, we use 5 runs and report means and 95\% confidence intervals. 
}
\label{table:appendix_optidice_and_baselines}
\begin{center}
\begin{small}
\begin{tabular}{l|rrrrrrrrr}
\toprule
  \multirow{2}{*}{Task} 
& \multirow{2}{*}{\shortstack[c]{BC}~} 
& \multirow{2}{*}{\shortstack[c]{SAC}} 
& \multirow{2}{*}{\shortstack[c]{BEAR}} 
& \multirow{2}{*}{\shortstack[c]{BRAC\\-v}} 
& \multirow{2}{*}{\shortstack[c]{Algae\\DICE}} 
& \multirow{2}{*}{\shortstack[c]{CQL}} 
& \multirow{2}{*}{\shortstack[c]{CQL\\(ours)}~~~} 
& \multicolumn{2}{c}{OptiDICE} \\
  &   &   &   &   &   & &   & \shortstack[c]{minimax}~ & \shortstack[c]{MSE}~~~~~ \\
\midrule
\midrule
maze2d-umaze              & 3.8   & 88.2 & 3.4  & -16.0 & -15.7             & 5.7 & -14.9 $\pm$ 0.7  & \textbf{111.0 $\pm$ 8.3}   & 105.8 $\pm$ 17.5  \\ 
maze2d-medium             & 30.3  & 26.1 & 29.0 & 33.8  & 10.0              & 5.0 & 17.2  $\pm$ 8.7  & 109.9 $\pm$ 7.7   & \textbf{145.2 $\pm$ 17.5}  \\ 
maze2d-large              & 5.0   & -1.9 & 4.6  & 40.6  & -0.1              & 12.5 & 1.6   $\pm$ 3.8  & 116.1 $\pm$ 43.1  & \textbf{155.7 $\pm$ 33.4}  \\ 
\hline
hopper-random             & 9.8   & 11.3 & 11.4 & \textbf{12.2}  & 0.9      & 10.8 & 10.7  $\pm$ 0.0  & 11.2  $\pm$ 0.1   & 10.7  $\pm$ 0.2   \\ 
hopper-medium             & 29.0  & 0.8  & 52.1 & 31.1  & 1.2               & 58.0 & 89.8  $\pm$ 7.6  & 92.9  $\pm$ 2.6   & \textbf{94.1  $\pm$ 3.7}   \\ 
hopper-medium-replay      & 11.8  & 3.5  & 33.7 & 0.6   & 1.1               & \textbf{48.6} & 33.3  $\pm$ 2.2  & 36.4  $\pm$ 1.1   & 30.7  $\pm$ 1.2   \\ 
hopper-medium-expert      & 111.9 & 1.6  & 96.3 & 0.8   & 1.1               & 98.7 & \textbf{112.3 $\pm$ 0.2}  & 111.5 $\pm$ 0.6   & 106.7 $\pm$ 1.8   \\ 
\hline
walker2d-random           & 1.6   & 4.1  & 7.3  & 1.9   & 0.5               & 7.0 &  3.0  $\pm$ 1.8  & 9.4   $\pm$ 2.2   & \textbf{9.9   $\pm$ 4.3}   \\ 
walker2d-medium           & 6.6   & 0.9  & 59.1 & \textbf{81.1}  & 0.3      & 79.2 & 73.7  $\pm$ 2.7  & 21.8  $\pm$ 7.1   & 20.8  $\pm$ 3.1   \\ 
walker2d-medium-replay    & 11.3  & 1.9  & 19.2 & 0.9   & 0.6               & \textbf{26.7} & 13.4  $\pm$ 0.8  & 21.5  $\pm$ 2.9   & 21.6  $\pm$ 2.1   \\ 
walker2d-medium-expert    & 6.4   & -0.1 & 40.1 & 81.6  & 0.4               & \textbf{111.0} & 99.7  $\pm$ 7.2  & 74.7  $\pm$ 7.5   & 74.8  $\pm$ 9.2   \\ 
\hline
halfcheetah-random        & 2.1   & 30.5 & 25.1 & 31.2  & -0.3     & \textbf{35.4} & 25.5  $\pm$ 0.5  & 8.3   $\pm$ 0.8   & 11.6  $\pm$ 1.2   \\ 
halfcheetah-medium        & 36.1  & -4.3 & 41.7 & \textbf{46.3}  & -2.2     & 44.4 & 42.3  $\pm$ 0.1  & 37.1  $\pm$ 0.1   & 38.2  $\pm$ 0.1   \\ 
halfcheetah-medium-replay & 38.4  & -2.4 & 38.6 & \textbf{47.7}  & -2.1     & 46.2 & 43.1  $\pm$ 0.7  & 38.9  $\pm$ 0.5   & 39.8  $\pm$ 0.3   \\ 
halfcheetah-medium-expert & 35.8  & 1.8  & 53.4 & 41.9  & -0.8              & 62.4 & 53.5  $\pm$ 13.3 & 76.2  $\pm$ 7.0   & \textbf{91.1  $\pm$ 3.7}   \\ 
\bottomrule
\end{tabular}
\end{small}
\end{center}
\vspace{-0.15in}
\end{table*}

~\newpage
\subsection{Experimental results with importance-weighted BC}

\begin{figure*}[h]
    \centering
	\includegraphics[width=1.0\textwidth]{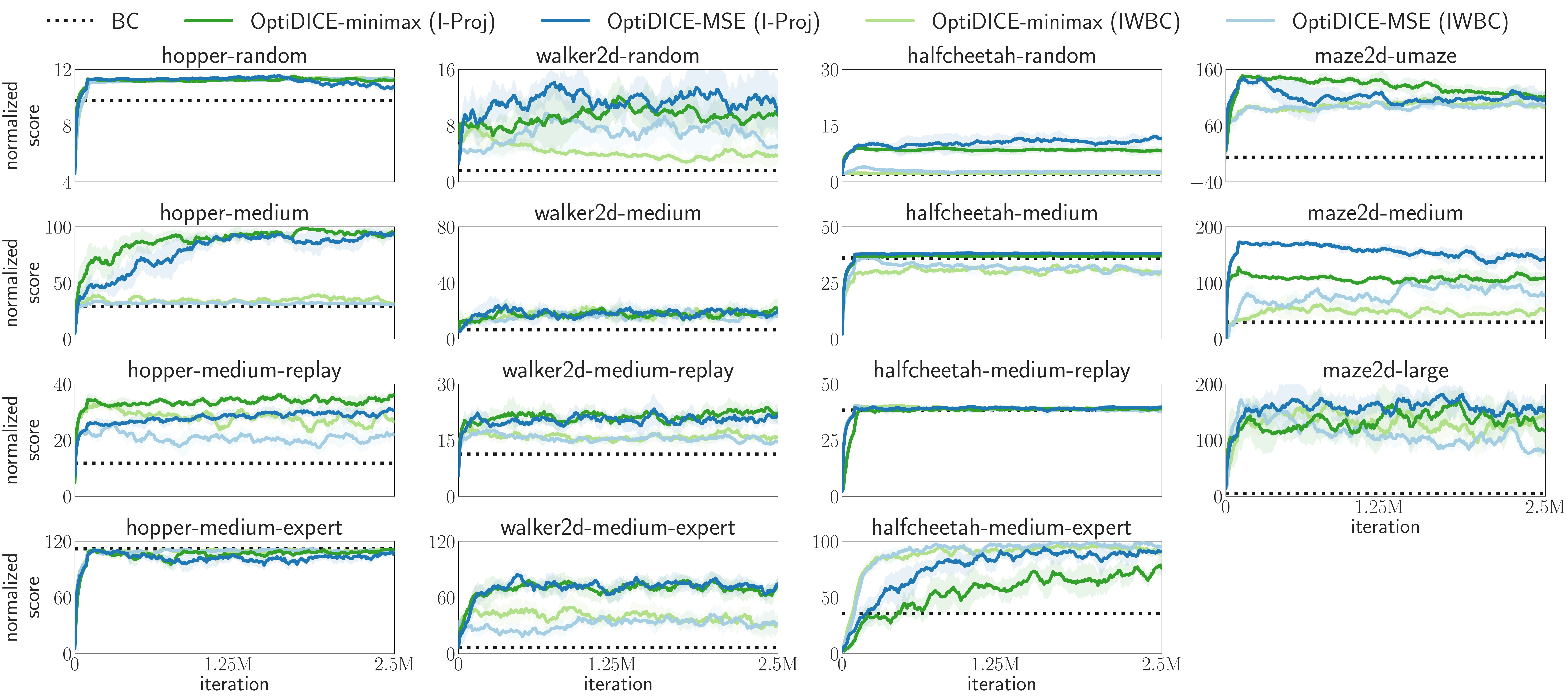}
	\vspace{-0.2in}
	\caption{Performance of BC, OptiDICE with importance-weighted BC (IWBC) and information projection (I-Proj) methods on D4RL benchmark. $\gamma=0.99$ is used. 
	}
	\vspace{-0.0in}
	\label{fig:figure_wbc}
\end{figure*}
The empirical results of OptiDICE for different policy extraction methods (information-weighted BC, I-projection methods) are depicted in \figureref{fig:figure_wbc}. 
For those results with IWBC, we search $\alpha$ within $\{0.0001, 0.001, 0.01, 0.1, 1\}$ and choose one with the best mean performance, which are summarized in \emph{Table}~\ref{table:iwbcparams}. The hyperparameters other than $\alpha$ are the same as those used for information-projection methods, which is described in Section~\ref{hyperparams}.
We empirically observe that policy extraction with information projection method performs better than the extraction with importance-weighted BC, as discussed in Section~\ref{subsection:policy_extraction}.

\begin{table}[h!]
\centering
\caption{Hyperaparameters for importance-weighted BC}
\label{table:iwbcparams}
\begin{tabular}{l | r | r }
\toprule
\multirow{1}{*}{Task} & \multicolumn{1}{c}{OptiDICE-MSE} & \multicolumn{1}{c}{OptiDICE-minimax} \\ \cline{2-3}
                                & $\alpha$      & $\alpha$      \\
\midrule
maze2d-umaze                    & 0.001         & 0.001         \\
maze2d-medium                   & 0.0001        & 0.001         \\
maze2d-large                    & 0.001         & 0.001          \\
\midrule
hopper-random                   & 1             & 1             \\
hopper-medium                   & 0.01          & 0.1           \\
hopper-medium-replay            & 0.1           & 0.1           \\
hopper-medium-expert            & 0.1           & 0.1             \\
\midrule
walker2d-random                 & 0.0001        & 0.0001        \\
walker2d-medium                 & 0.1           & 0.1           \\
walker2d-medium-replay          & 0.1           & 0.1           \\
walker2d-medium-expert          & 0.01          & 0.01          \\
\midrule
halfcheetah-random              & 0.001         & 0.01          \\
halfcheetah-medium              & 0.0001        & 0.1           \\
halfcheetah-medium-replay       & 0.01          & 0.01          \\
halfcheetah-medium-expert       & 0.01          & 0.01          \\
\bottomrule
\end{tabular}
\end{table}

~\newpage
\subsection{Experimental results for $\gamma \in \{0.99, 0.999, 0.9999, 1.0\}$}

\begin{figure*}[h]
    \centering
	\includegraphics[width=1.0\textwidth]{images/main_figure.pdf}
	\vspace{-0.2in}
	\caption{Performance of BC, CQL, 
	OptiDICE-minimax and OptiDICE-MSE on D4RL benchmark for $\gamma=0.99$. 
	}
	\vspace{0.4in}
	\includegraphics[width=1.0\textwidth]{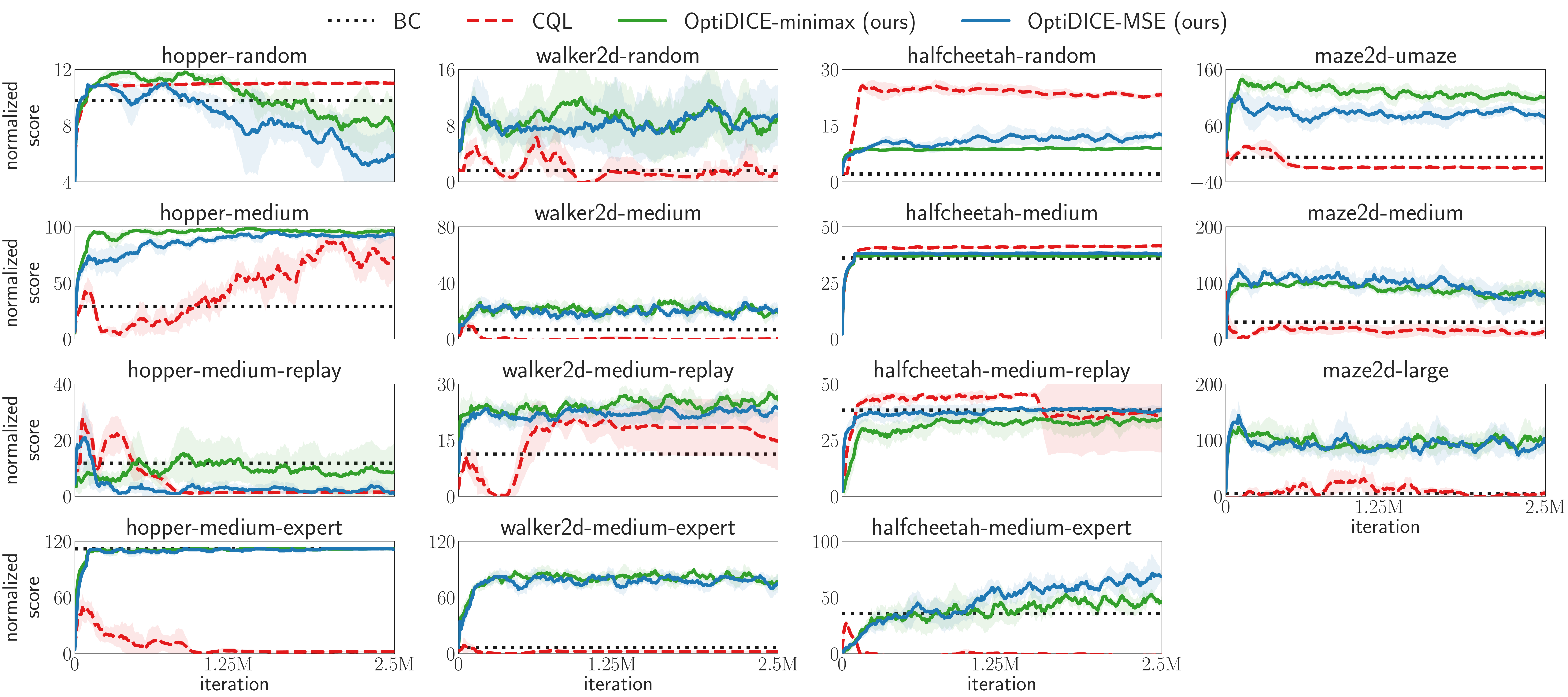}
	\vspace{-0.2in}
	\caption{Performance of BC, CQL, 
	OptiDICE-minimax and OptiDICE-MSE on D4RL benchmark for $\gamma=0.999$. 
	}
\end{figure*}
~
\begin{figure*}[h]
    \centering
	\includegraphics[width=1.0\textwidth]{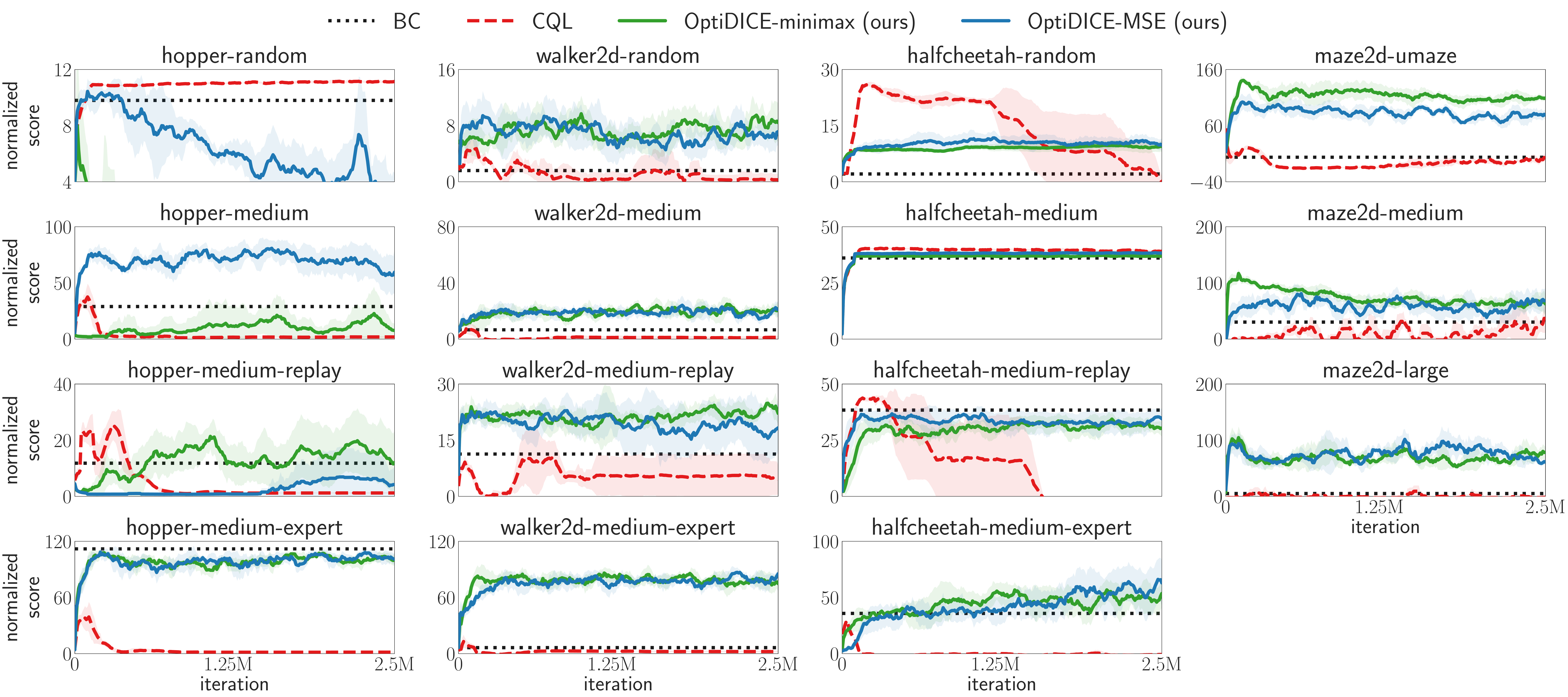}
	\vspace{-0.2in}
	\caption{Performance of BC, CQL, 
	OptiDICE-minimax and OptiDICE-MSE on D4RL benchmark for $\gamma=0.9999$. 
	}
	\vspace{0.4in}
	\includegraphics[width=1.0\textwidth]{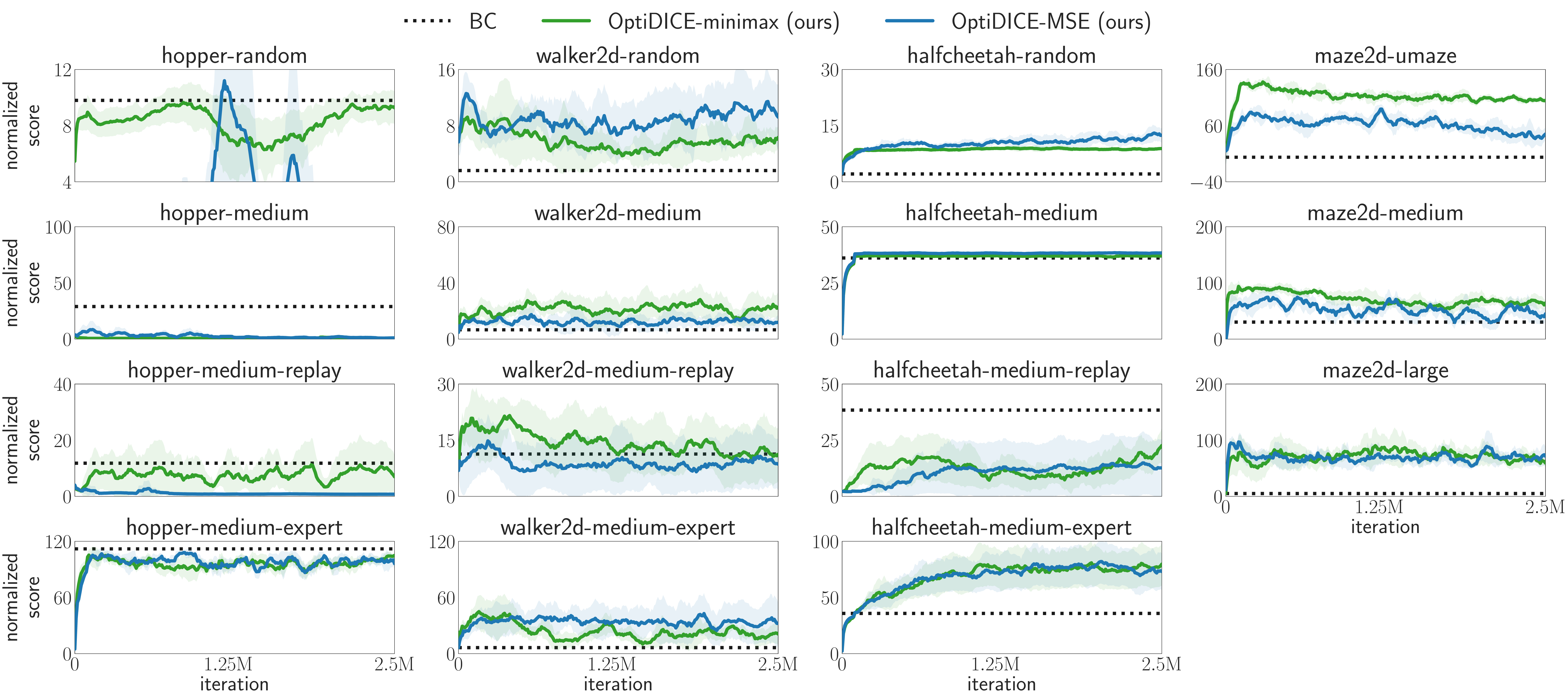}
	\vspace{-0.2in}
	\caption{Performance of BC,
	OptiDICE-minimax and OptiDICE-MSE on D4RL benchmark for $\gamma=1$. Note that CQL cannot deal with $\gamma=1$ case. Thus, we only provide the results of OptiDICE and BC.
	}
\end{figure*}

\end{document}